%% file: acl.tex
\pdfoutput=1

\documentclass[11pt]{article}

\usepackage[final]{acl}

\usepackage{times}
\usepackage{latexsym}

\usepackage[T1]{fontenc}

\usepackage[utf8]{inputenc}

\usepackage{microtype}

\usepackage{graphicx}

\usepackage{amsfonts}
\usepackage{amsmath}

\usepackage[labelformat=simple]{subcaption}

\usepackage{stackengine}

\usepackage{adjustbox}
\usepackage{booktabs}
\usepackage{multirow}

\usepackage{nameref}

\title{Anticipation-Free Training for Simultaneous Machine Translation}

\author {
    Chih-Chiang Chang,\quad
    Shun-Po Chuang,\quad
    Hung-yi Lee \\
    National Taiwan University \\
    \texttt{\{r09922057,f04942141,hungyilee\}@ntu.edu.tw}
}

\begin{document}
\maketitle
\begin{abstract}
\input{sections/abstract}
\end{abstract}

\input{sections/introduction}
\input{sections/related}
\input{sections/proposed}
\input{sections/exp}
\input{sections/quant_exp}
\input{sections/qual_exp}
\input{sections/ablation}

\input{sections/conclusion}


\bibliography{refs}
\bibliographystyle{acl_natbib}

\appendix
\input{sections/appendix}

\end{document}

%% file: sections/abstract.tex
Simultaneous machine translation (SimulMT) speeds up the translation process by starting to translate before the source sentence is completely available. It is difficult due to limited context and word order difference between languages.  
Existing methods increase latency or introduce adaptive read-write policies for SimulMT models to handle local reordering and improve translation quality.
However, the long-distance reordering would make the SimulMT models learn translation mistakenly. 
Specifically, the model may be forced to predict target tokens when the corresponding source tokens have not been read. 
This leads to aggressive anticipation during inference, resulting in the hallucination phenomenon. To mitigate this problem, we propose a new framework that decompose the translation process into the monotonic translation step and the reordering step, and we model the latter by the auxiliary sorting network (ASN). 
The ASN rearranges the hidden states to match the order in the target language, so that the SimulMT model could learn to translate more reasonably. 
The entire model is optimized end-to-end and does not rely on external aligners or data. 
During inference, ASN is removed to achieve streaming. 
Experiments show the proposed framework could outperform previous methods with less latency.

%% file: sections/introduction.tex
\section{Introduction}
\label{sec:intro}

\begin{figure}[t!]
    \centering
    \includegraphics[width=0.95\linewidth]{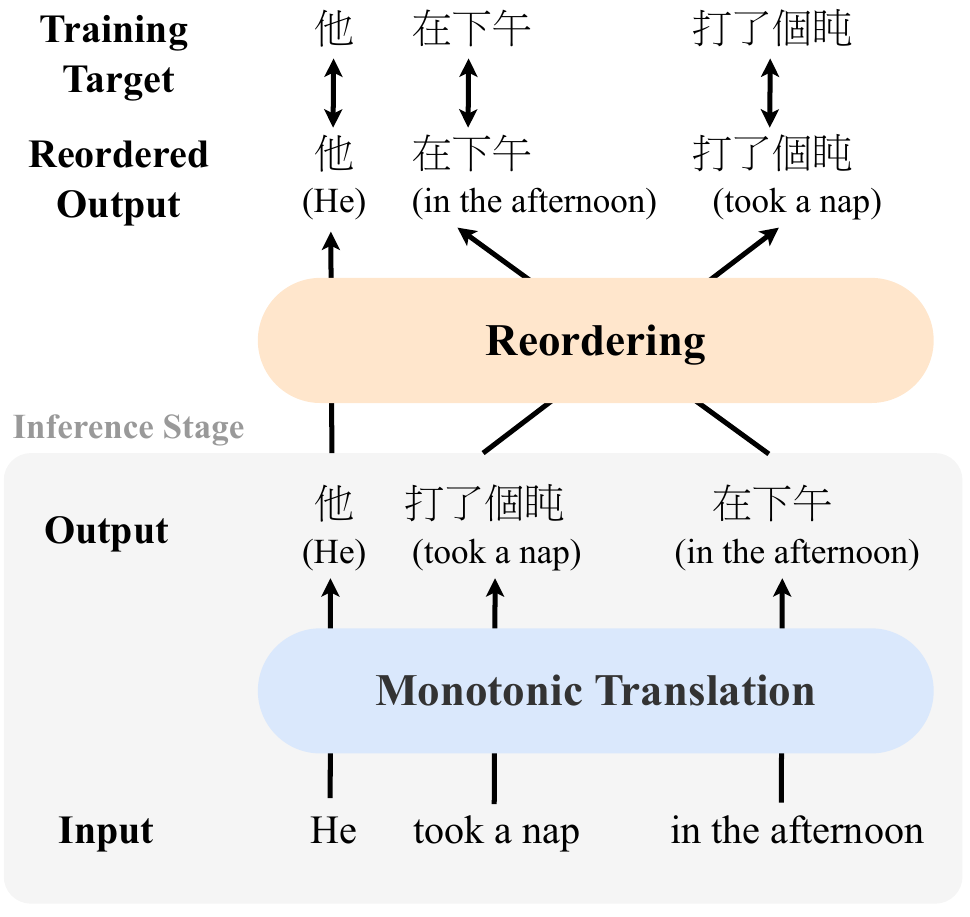}
    
    \caption{Illustration of the training process. The translated output is rearranged to match the order of  training target, reducing anticipation. We use the gray part during inference.}
    \label{fig:overview}
\end{figure}

Simultaneous machine translation (SimulMT) is an extension of neural machine translation (NMT), aiming to perform streaming translation by outputting the translation before the source input has ended. It is more applicable to real-world scenarios such as international conferences, where people could communicate fluently without delay. 

However, SimulMT faces additional difficulties compared to full-sentence translation -- such a model needs to translate with limited context, and the different word order between languages would make streaming models learn translation mistakenly. 
The problems can often be alleviated by increasing the context. 
Using more context allows the model to translate with more information,
trading off speed for quality.
But the word order could be very different among languages.
Increasing the context could only solve the local reordering problem. 
If long-distance reordering exists in training data, 
the model would be forced to predict tokens in the target language when the corresponding source tokens have not been read. this is called \textit{anticipation}~\cite{ma2019stacl}.
Ignoring the long-distance reordering may cause unnecessarily high latency, or encourage aggressive anticipation, resulting in the hallucination phenomenon~\cite{muller2020domain}.

It sheds light on the importance of matching the word order between the source and target languages. 
Existing methods aim to reduce anticipation
by using syntax-based rules to rewrite the translation target~\cite{he-etal-2015-syntax}.
It requires additional language-specific prior knowledge and constituent parse trees.
Other approaches pre-train a full-sentence model, then incrementally feed the source sentence to it to generate monotonic translation target (pseudo reference)~\cite{chen2021improving, zhang2020learning}.
However, the full-sentence model was not trained to translate incrementally, which creates a train-test mismatch, resulting in varying prediction quality. They require combining with the original data to be effective.


To this end, this work aims to address long-distance reordering by incorporating it directly into the training process, as Figure~\ref{fig:overview} shows.
We decompose the typical translation process into the \textit{monotonic translation} step and the \textit{reordering} step. 
Inspired by the Gumbel-Sinkhorn network~\cite{mena2018learning}, we proposed an auxiliary sorting network (ASN) for the \textit{reordering} step. 
During training, the ASN explicitly rearranges the hidden states to match the target language word order. 
The ASN will not be used during inference, so that the model could translate monotonically. The proposed method reduces anticipation, thus increases the lexical precision~\cite{he-etal-2015-syntax} of the model without compromising its speed. 
We apply the proposed framework to a simple model -- a causal Transformer encoder trained with connectionist temporal classification (CTC)~\cite{graves2006connectionist}. The CTC loss can learn an adaptive policy~\cite{chousa2019simultaneous}, which performs local reordering by predicting blank symbols until enough information is read, then write the information in the target order. Even so, it still suffers from high latency and under-translation due to long-distance reordering in training data. Our ASN handles these long-distance reordering, improving both the latency and the quality of the CTC model. 
We conduct experiments on CWMT English to Chinese and WMT15 German to English translation datasets.
Our contributions are summarized below:
\begin{itemize}
    \setlength\itemsep{0em}
    \item We proposed a new framework for SimulMT. The ASN could apply on various causal models to handle long-distance reordering.
    \item Experiments showed that the proposed method could outperform the pseudo reference method. It indicated the proposed method could better handle the long-distance reordering.
    \item The proposed model is a causal encoder, which is parameter efficient and could outperform wait-$k$ Transformer with less latency.
\end{itemize}
Our implementation is based on fairseq~\cite{ott-etal-2019-fairseq}. 
The instructions to access our source code is provided in Appendix~\ref{appendix:code}.

%% file: sections/related.tex
\section{Related Works}
\label{sec:related}

\subsection{Simultaneous Translation}
\label{subsec:related_simul}

SimulMT is first achieved by applying fixed read-write policies on NMT models.
Wait-if-worse and Wait-if-diff~\cite{cho2016can} form decisions based on the next prediction's probability or its value.
Static Read and Write~\cite{dalvi2018incremental} first read several tokens, then repeatedly read and write several tokens at a time. Wait-$k$~\cite{ma2019stacl} trains end-to-end models for SimulMT. Its policy is similar to Static Read and Write.

On the other hand, adaptive policies seek to learn the read-write decisions. Some works explored training agents with reinforcement learning (RL)~\cite{gu2017learning, luo2017learning}. Others design expert policies and apply imitation learning (IL)~\cite{zheng2019simpler,zheng2019simultaneous}. Monotonic attention~\cite{raffel2017online} integrates the read-write policy into the attention mechanism to jointly train with NMT. MoChA~\cite{chiu2018monotonic} enhances monotonic attention by adding soft attention over a small window. 
MILk~\cite{arivazhagan2019monotonic} extends such window to the full encoder history. 
MMA~\cite{ma2019monotonic} extends MILk to multi-head attention.
Connectionist temporal classification (CTC) were also explored for adaptive policy by treating the blank symbol as wait action~\cite{chousa2019simultaneous}. 
Recently, making read-write decisions based on segments of meaningful unit (MU)~\cite{zhang2020learning} improves the translation quality.
Besides, an adaptive policy can also be derived from an ensemble of fixed-policy models~\cite{zheng-etal-2020-simultaneous}.

When performing simultaneous interpretation, humans avoid long-distance reordering whenever possible~\cite{al2000use, he-etal-2016-interpretese}. Thus, 
some works seek to reduce the anticipation in data to ease the training of simultaneous models.
These include syntax-based rewriting~\cite{he-etal-2015-syntax},
or generating pseudo reference by
test-time wait-$k$~\cite{chen2021improving}
and prefix-attention~\cite{zhang2020learning}. 
We reduce anticipation from a different approach: instead of rewriting the target, we let the model match its hidden states to the target on its own. As shown in experiments, our method is comparable or superior to the pseudo reference method.

\subsection{Gumbel-Sinkhorn Network}
\label{subsec:related_gumbel}


The Sinkhorn Normalization~\cite{adams2011ranking} is an iterative procedure that converts a matrix into doubly stochastic form. It was initially proposed to perform gradient-based rank learning. Gumbel-Sinkhorn Network~\cite{mena2018learning} combines the Sinkhorn Normalization with the Gumbel reparametrization trick~\cite{kingma2013auto}. It approximates sampling from a distribution of permutation matrices. Subsequently, Sinkhorn Transformer~\cite{tay2020sparse} applied this method to the Transformer~\cite{vaswani2017attention} 
to model long-distance dependency in language models with better memory efficiency.
This work applies the Gumbel-Sinkhorn Network to model the reordering between languages, in order to reduce anticipation in SimulMT.

%% file: sections/proposed.tex
\section{Proposed Method}
\newcommand{\rma}{\boldsymbol{\mathrm{A}}}
\newcommand{\rmx}{\boldsymbol{\mathrm{x}}}
\newcommand{\rmy}{\boldsymbol{\mathrm{y}}}
\newcommand{\rmz}{\boldsymbol{\mathrm{Z}}}
\newcommand{\rmh}{\boldsymbol{\mathrm{H}}}
\newcommand{\rmq}{\boldsymbol{\mathrm{Q}}}
\newcommand{\rmi}{\boldsymbol{\mathrm{I}}}
\newcommand{\rmeps}{\boldsymbol{\mathcal{E}}}
\newcommand{\colvecone}{\boldsymbol{\mathrm{1}}}
\newcommand{\rulesep}{\unskip\ \vrule\ }

\begin{figure*}[ht]
    \centering
    \def\figa{\includegraphics[width=0.6\linewidth]{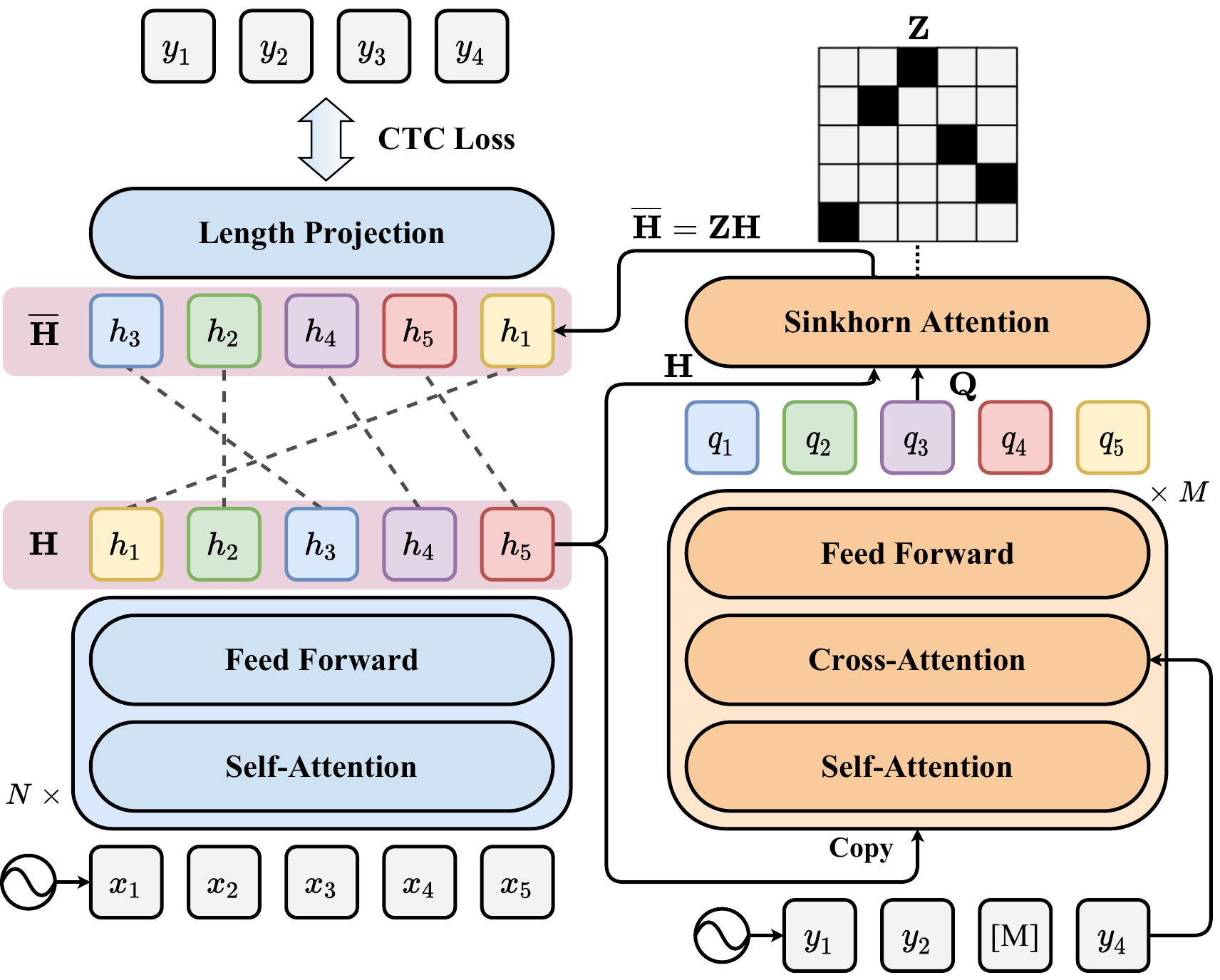}} 
    \def\figb{\includegraphics[width=0.29\linewidth]{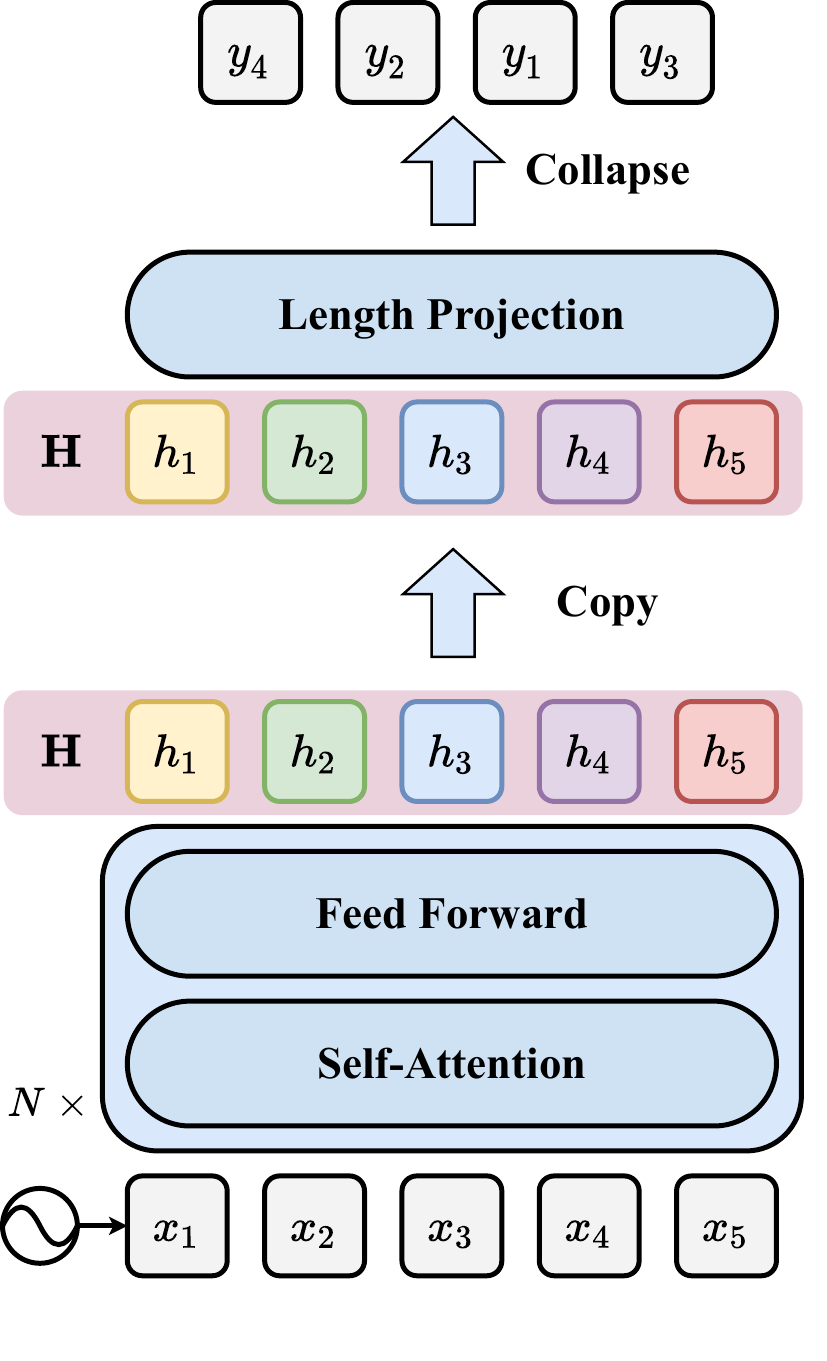}} 

    \def\capa{The model consists of a causal encoder (lower left, blue), an ASN (right, orange), and a length projection network (upper left, blue). ``[M]'' is the masking embedding.}
    \def\capb{During inference, only the encoder and length projection (blue) are used.}
    
    \subcaptionbox{\capa\label{fig:arch_overall}}{\figa}
    \hspace{2ex}
    \subcaptionbox{\capb\label{fig:arch_inference}}{\figb}
    
    \caption{The architecture of the proposed model. Add \& Norm layers are omitted for simplicity.}
    \label{fig:architecture}
\end{figure*}


For a source sentence $\rmx=\langle x_1, x_2, ... , x_{|\rmx|} \rangle$ and a target sentence $\rmy=\langle y_1, y_2, ... , y_{|\rmy|} \rangle$, in order to perform SimulMT, the conditional probability of translation $p(\rmy|\rmx)$ is modeled by the prefix-to-prefix framework~\cite{ma2019stacl}. Formally,
\begin{equation}
    p_g(\rmy|\rmx) = \prod_{t=1}^{|\rmy|} p(y_{t}|\rmx_{\leq g(t)}, \rmy_{<t}).
\end{equation}
where $g(t)$ is a monotonic non-decreasing function. This way, the $t$-th token $\hat{\rmy}_{t}$ can be predicted with a limited context $\rmx_{\leq g(t)}$.
However, if long-distance reordering exists in the training data, the model is forced to generate target tokens whose corresponding source tokens have not been revealed yet. This issue is known as anticipation.
\par

\subsection{Training Framework}
To overcome this, we introduce a latent variable $\rmz$: a permutation matrix capturing the reordering process from $\rmx$ to $\rmy$. Thus, the translation probability can be expressed as a marginalization over $\rmz$:
\begin{equation}
    p(\rmy|\rmx) = \sum_{\rmz}
    \underbrace{p_g(\rmy|\rmx, \rmz)}_{\substack{\text{monotonic} \\ \text{translation}}}
    \underbrace{p(\rmz|\rmx)}_{\text{reordering}}.
    \label{eq:marginalize_conditional}
\end{equation}
During training, since $\rmz$ captures reordering, the $p_g(\rmy|\rmx, \rmz)$ corresponds to monotonic translation, which can be correctly modeled by a prefix-to-prefix model without anticipation. During inference, we can translate monotonically by simply removing the effect of $\rmz$:
\begin{equation}
    \hat{\rmy} = \arg\max_{\rmy} p_g(\rmy|\rmx, \rmz=\rmi).
    \label{eq:remove_z}
\end{equation}
where $\rmi$ is the identity matrix. However, equation~\ref{eq:marginalize_conditional} is intractable due to the factorial search space of permutations. 
One could select the most likely permutation using an external aligner~\cite{Ran_Lin_Li_Zhou_2021}, but such a method requires an external tool, and it could not be end-to-end optimized. 
Instead, we use the ASN to learn the permutation matrix $\rmz$ associated with source-target reordering. By doing this, the entire model is optimized end-to-end.
\par
Figure~\ref{fig:architecture} shows the proposed framework applied on the CTC model. It is composed of a causal Transformer encoder, an ASN, and a length projection network. We describe each component in detail below. 

\subsection{Causal Encoder}
The encoder maps the source sequence $\rmx$ to hidden states $\rmh=\langle h_1, h_2, ... , h_{|\rmx|} \rangle$. During training, the encoder uses a causal attention mask so that it can be streamed during inference. To enable the trade-off between quality and latency, we introduce a tunable delay in the causal attention mask of the first encoder layer. We define the delay in a similar sense to wait-$k$: For delay-$k$, the $t$-th hidden state $h_t$ is computed after observing the $(t+k-1)$-th source token. 
\par
We pre-train the encoder with CTC loss~\cite{libovicky2018end}.
Since the CTC is an adaptive policy already capable of local reordering, initializing from it encourages the ASN to only handle long-distance reordering. We study the effectiveness of this technique in Section~\ref{sec:ablation-weight-init}.



\subsection{Auxiliary Sorting Network (ASN)}\label{sec:asn}
The ASN samples a permutation matrix $\rmz$, which would sort the encoder hidden states $\rmh$ into the target order. To do so, the ASN first computes intermediate variables $\rmq=\langle q_1, q_2, ... , q_{|\rmx|} \rangle$ using a stack of $M$ non-causal Transformer decoder layers. 
These layers use the target token embeddings as the context for cross attention. 
Providing this context guides the reordering process\footnote{Although ASN has decoder layers and takes target tokens as input, which are unavailable during inference, they are only used to assist training.}, 
inspired by the word alignment task~\cite{zhang-van-genabith-2021-bidirectional,chen-etal-2021-mask}. 
We randomly mask out $\gamma\%$ of the context in ASN to avoid collapsing to a trivial solution.

Subsequently, the Sinkhorn Attention in ASN computes the attention scores between $\rmq$ and $\rmh$ using the scaled dot-product attention:
\begin{equation}
    \rma = \frac{\rmq\rmh^{T}}{\sqrt{d_h}}, \label{eq:attn2} 
\end{equation}
where $d_h$ is the last dimension of $\rmh$.
To convert the attention scores $\rma$ to a permutation matrix $\rmz$, ASN applies the Gumbel-Sinkhorn operator. Such operator approximates sampling from a distribution of permutation matrices~\cite{mena2018learning}.
It is described by first adding the Gumbel noise (equation~\ref{eq:samplegumbel}), then scaling by a positive temperature $\tau$, and finally applying the $l$-iteration Sinkhorn normalization (denoted by $S^{l}(\cdot)$)~\cite{adams2011ranking}. We also add a scaling factor $\delta$ to adjust the Gumbel noise level (equation~\ref{eq:deltagumbel}). 
The output would be doubly stochastic~\cite{sinkhorn1964relationship}, which is a relaxation of permutation matrix. 
We leave the detailed description of the Gumbel-Sinkhorn operator in Appendix~\ref{appendix:gumbel_sinkhorn}.
\begin{align}
    \rmeps &\in\mathbb{R}^{N\times N}  \overset{i.i.d.}{\sim} Gumbel(0, 1), \label{eq:samplegumbel} \\
    \rmz &= S^{l}\left(\left(\rma+\delta\rmeps\right)/\tau\right), \label{eq:deltagumbel} 
\end{align}

Next, we use a matrix multiplication of $\rmz$ and $\rmh$ to reorder $\rmh$, the result is denoted by $\overline{\rmh}$: 
\begin{equation}
    \overline{\rmh} = \rmz \rmh \label{eq:matmul}
\end{equation}

Since $\rmz$ approximates a permutation matrix, using matrix multiplication is equivalent to permuting the vectors in $\rmh$. This preserves the content of its individual vectors, and is essential to our method as we will show in Section~\ref{sec:ablation-gumbel}.

\subsection{Length Projection}

To optimize the model with CTC loss function, we tackle the length mismatch between $\overline{\rmh}$ and $\rmy$ by projecting $\overline{\rmh}$ to a $\mu$-times longer sequence via an affine transformation~\cite{libovicky2018end}.
The $\mu$ represents the upsample ratio.
For ASN to learn reordering effectively, it is required that the projection network and the loss must not perform reordering. Our length projection is time-independent, and CTC is monotonic, both satisfy our requirement.


\subsection{Inference Strategy}
To enable streaming, we remove the ASN during inference\footnote{While this seemingly creates a train-test discrepancy, we address this in~\nameref{appendix:faq}} (Figure~\ref{fig:arch_inference}). Specifically, when a new input token $x_t$ arrives, the encoder computes the hidden state $h_t$, then we feed $h_t$ directly to the length projection to predict the next token(s). 
The prediction is post-processed by the CTC collapse function in an online fashion. Namely, we only output a new token if 1) it is not the blank symbol and 2) it is different from the previous token.

%% file: sections/exp.tex
\section{Experiments}
\label{sec:exp}
\subsection{Datasets}
We conduct experiments on English-Chinese and German-English datasets.
For En-Zh,
we use a subset\footnote{We use casia2015, casict2011, casict2015, neu2017.} of CWMT~\cite{chen2019machine} parallel corpora as training data (7M pairs). We use NJU-newsdev2018 as the development set and report results on CWMT2008, CWMT2009, and CWMT2011. The CWMT test sets have up to 3 references. Thus we report the 3-reference BLEU score.
For De-En,
we use WMT15~\cite{bojar-EtAl:2015:WMT} parallel corpora as training data (4.5M pairs). We use newstest2013 as the development set and report results on newstest2015.
\par
We use SentencePiece~\cite{kudo2018sentencepiece} on each language separately to obtain its vocabulary of 32K subword units. We filter out sentence pairs that have empty sentences or exceed 1024 tokens in length.

\subsection{Experimental Setup}
All SimulMT models use causal encoders. During inference, the encoder states are computed incrementally after each read, similar to~\cite{Elbayad2020}.
The causal encoder models follow a similar training process to non-autoregressive translation (NAT)~\cite{gu2018non,libovicky2018end,lee2018deterministic,zhou2019understanding}.
We adopt sequence level knowledge distillation (Seq-KD)~\cite{kim2016sequence} for all systems.
The combination of Seq-KD and CTC loss has been shown to achieve state-of-the-art performance~\cite{gu2020fully} and could deal with the reordering problem~\cite{chuang-etal-2021-investigating}.
Specifically, we first train a full-sentence model as a teacher model on the original dataset, then we use beam search with beam width $5$ to decode the Seq-KD set. We use the Seq-KD set in subsequent experiments. 
We list the Transformer and ASN hyperparameters separately in Appendix~\ref{appendix:hyper} and~\ref{appendix:asn_hyper}.

\par
We use Adam~\cite{kingma2014adam} with an inverse square root schedule for the optimizer. The max learning rate is 5e-4 with 4000 warm-up steps. We use gradient accumulation to achieve an effective batch size of 128K tokens for the teacher model and 32K for others. 
We optimize the model with the 300K steps. Early stopping is applied when the validation BLEU does not improve within 25K steps.
Label smoothing~\cite{szegedy2016rethinking} with $\epsilon_{ls}=0.1$ is applied on cross-entropy and CTC loss. For CTC, this reduces excessive blank symbol predictions~\cite{suyoun2017improved}. Random seeds are set in training scripts in our source code. For the hardware information and environment settings, see Appendix~\ref{appendix:hardware}.
\par
For latency evaluation, we use SimulEval~\cite{ma2020simuleval} to compute Average Lagging (AL)~\cite{ma2019stacl} and Computation Aware Average Lagging (AL-CA)~\cite{ma2020simulmt}. AL is measured in words or characters, whereas AL-CA is measured in milliseconds.
We describe these metrics in detail in Appendix~\ref{appendix:detail_metrics}.
For quality evaluation, we use BLEU~\cite{papineni2002bleu} calculated by  SacreBLEU~\cite{post-2018-call}.
We conduct statistical significance test for BLEU using paired bootstrap resampling~\cite{koehn2004statistical}.
For multiple references, we use the first reference to run SimulEval\footnote{we use SimulEval for latency metrics only. Only one reference is required to run it.} and use all available references to run SacreBLEU. The language-specific settings for SimulEval and SacreBLEU can respectively be found in Appendix~\ref{appendix:simuleval_config} and~\ref{appendix:signature}.

\subsection{Baselines}
We compare our method with two target rewrite methods which generate new datasets:
\begin{itemize}
    \setlength\itemsep{0em}
    \item \textbf{Pseudo reference}~\cite{chen2021improving}: This approach first trains a full-sentence model and uses it to generate monotonic translation. The approach applies the test-time wait-$k$ policy~\cite{ma2019stacl}, and performs beam search with beam width $5$ to generate pseudo references. The pseudo reference set is the combination of original dataset and the pseudo references.
    We made a few changes 1) instead of the full-sentence model, we use the wait-9 model\footnote{our wait-9 model has higher training set BLEU score than applying test-time wait-$k$ on full-sentence model.}. 2) instead of creating a new dataset for each $k$, we only use $k=9$ since it has the best quality. 
    \item \textbf{Reorder}: We use the word alignments to reorder the target sequence. We use \textit{awesome-align}~\cite{dou2021word} to obtain word alignments on the Seq-KD set, and we sort the target tokens based on their corresponding source tokens. Target tokens that did not align to a source token are placed at the position after their preceding target token.
\end{itemize}
We train two types of models on either the Seq-KD set, the pseudo reference set or the reorder set:
\begin{itemize}
    \setlength\itemsep{0em}
    \item \textbf{wait-$k$}: an encoder-decoder model. It uses a fixed policy that first reads $k$ tokens, then repeatedly reads and writes a single token.
    \item \textbf{CTC}: a causal encoder trained with CTC loss. The policy is adaptive, i.e., it outputs blank symbols until enough content is read, outputs the translated tokens, then repeats.
\end{itemize}


%% file: sections/quant_exp.tex
\subsection{Quantitative Results}
\label{sec:quantitative}

\begin{figure*}[h]
    \centering
    \def\figwid{0.49\textwidth}
    \includegraphics[width=\figwid]{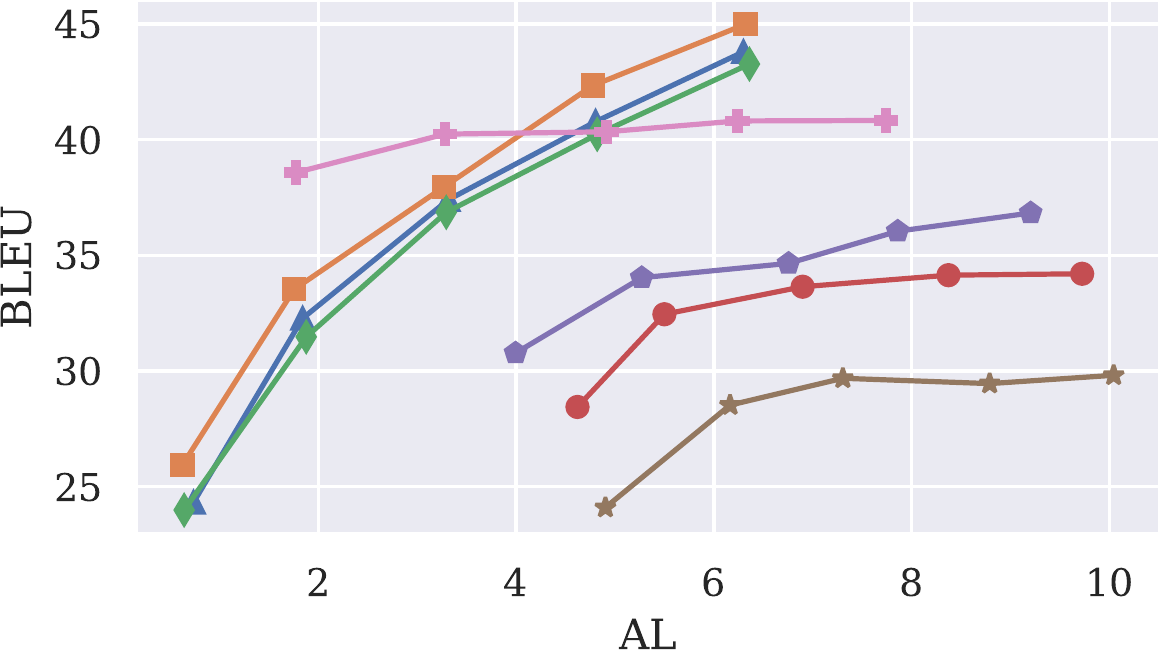}
    \includegraphics[width=\figwid]{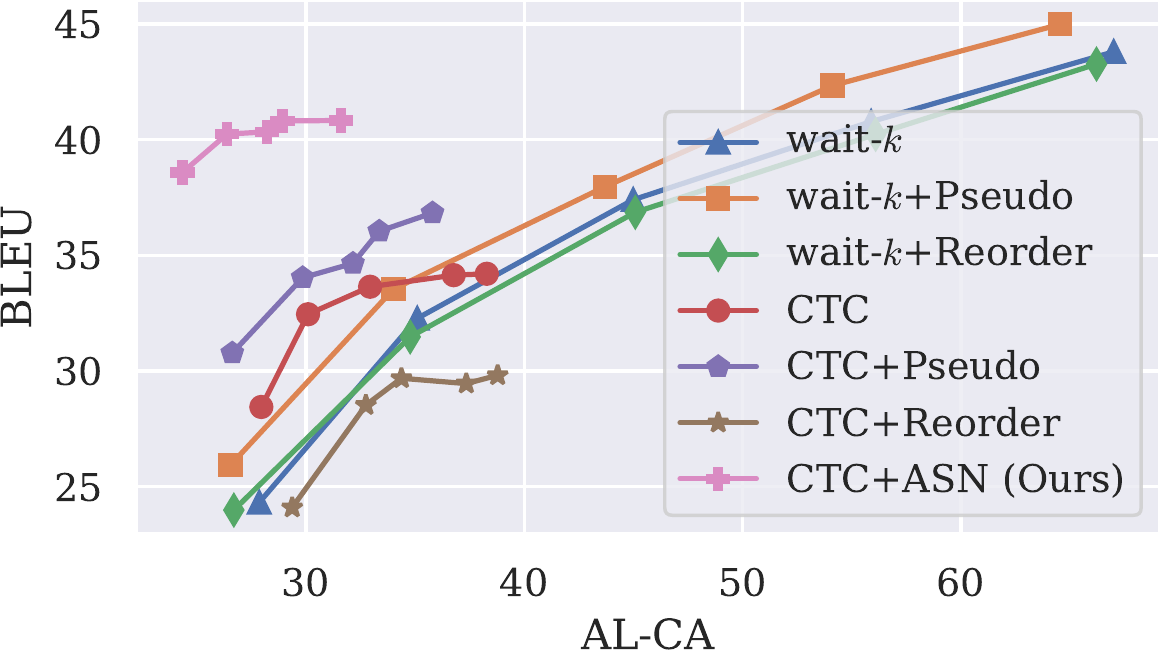}
    
    \caption{Latency-quality trade off on the CWMT En-Zh dataset. Each line represents a system, and the 5 nodes from left to right corresponds to $k=1,3,5,7,9$. The figures share the legend.}
    \label{fig:latency-quality-enzh}
\end{figure*}

    

\begin{figure}[ht]
    \centering
    \def\figwid{0.49\textwidth}
    \includegraphics[width=\figwid]{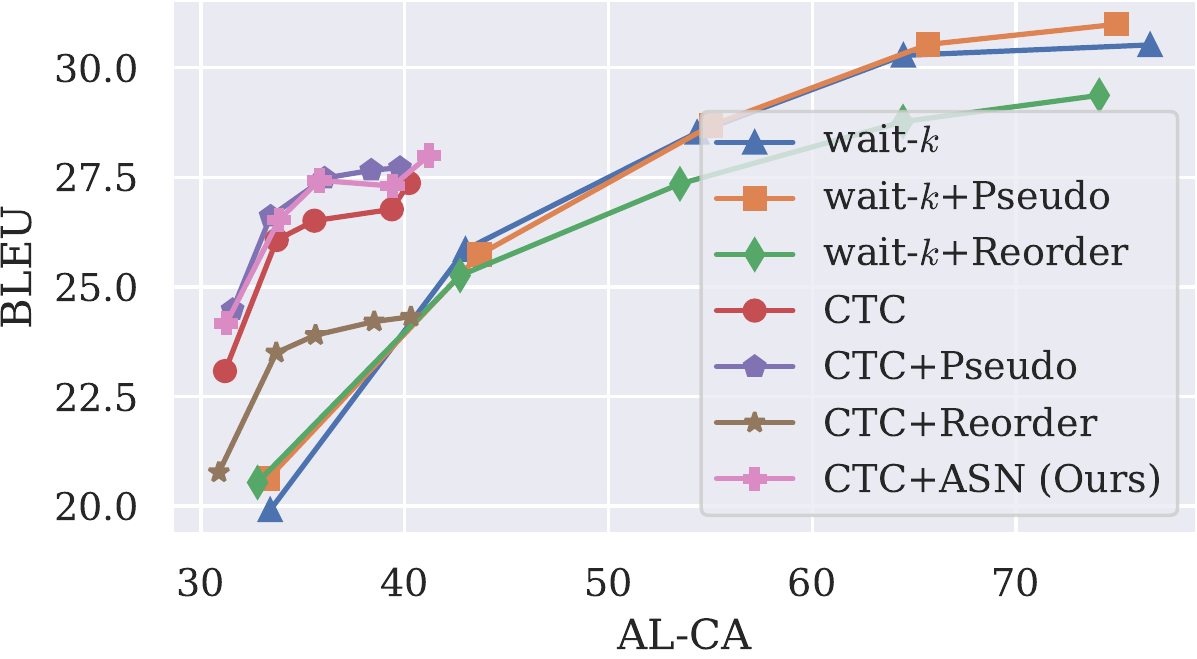}
    
    \caption{Latency-quality trade off on the WMT15 De-En dataset.}
    \label{fig:latency-quality-deen}
\end{figure}

Figure~\ref{fig:latency-quality-enzh} shows the latency-quality trade-off on the CWMT dataset, each node on a line represents a different value of $k$. 
Due to space limit, the significant test results are reported in Appendix~\ref{appendix:detailed_results}.


First of all, although the vanilla CTC model has high latency in terms of AL, they are comparable to or faster than the wait-$k$ model according to AL-CA. This is due to the reduced parameter size. Besides, CTC models outperform wait-$k$ in low latency settings.
The pseudo reference method improves the quality of wait-$k$ and CTC models, and it slightly improves the latency of the CTC model. In contrast, the reorder method harms the performance of both models.
Meanwhile, our method significantly improves both the quality and latency of the CTC model across all latency settings, outperforming the pseudo reference method and the reorder method. In particular, our $k=1,3$ models outperform wait-1 by around 13-15 BLEUs with a faster speed in terms of AL-CA. This shows that our models are more efficient than wait-$k$ models under low latency regimes.
\par
Figure~\ref{fig:latency-quality-deen} shows the latency-quality trade-off on the WMT15 De-En dataset. The vanilla CTC model is much more competitive in De-En. It outperforms vanilla wait-k in low latency settings in BLEU and AL-CA, and its AL is much less than those in En-Zh. 
Our method improves the quality of the CTC model, comparable to the pseudo reference method. However, our method does not require combining with the original dataset to improve the performance.

\par
\begin{figure}[h]
    \centering
    \def\figwid{0.95\linewidth}
    \includegraphics[width=\figwid]{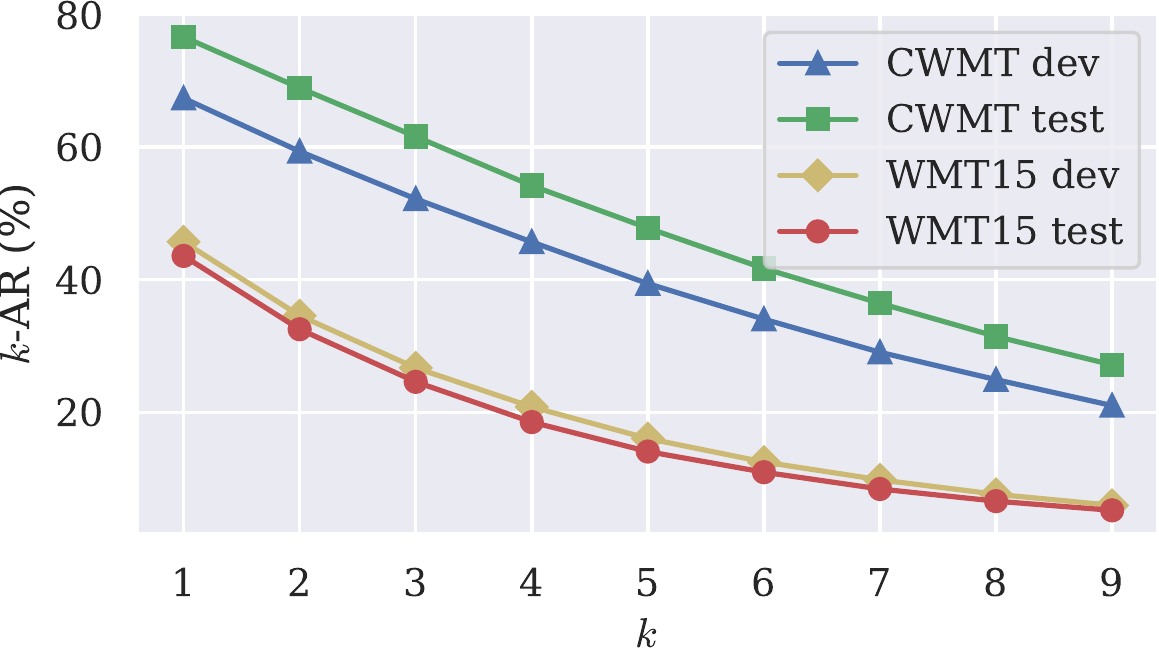}
    \caption{The $k$-anticipation rate computed on CWMT En-Zh and WMT15 De-En development and test sets.}
    \label{fig:kar}
\end{figure}

To understand why our method is more effective on CWMT, we calculate the $k$-Anticipation Rate ($k$-AR)~\cite{chen2021improving} on the evaluation sets of both datasets. 
For the definition of $k$-AR, see Appendix~\ref{appendix:detail_metrics}.
Intuitively, $k$-AR describes the amount of anticipation (or reordering) in the corpus whose range is longer than $k$ source tokens.
We report $k$-AR across $1\leq k\leq9$ in Figure~\ref{fig:kar}. 
En-Zh has much higher $k$-AR in general, and it decreases slower as $k$ increases. When $k=9$, over 20\% of anticipations remain in En-Zh, while almost none remains in De-En. We conclude that En-Zh has much more reordering, and over 20\% of them are longer than 9 words. The abundance of long-distance reordering gives our method an advantage, which explains the big improvement observed on CWMT. On the other hand, De-En reordering is less common and mostly local, so ASN has limited effect. Indeed, we found that ASN predicts matrices close to the identity matrix on De-En, whereas, on En-Zh, it predicts non-identity matrices throughout training.

%% file: sections/qual_exp.tex
\subsection{Qualitative Results}
\begin{figure*}[th]
    \centering
    \def\figwid{\textwidth}
    \includegraphics[width=\figwid]{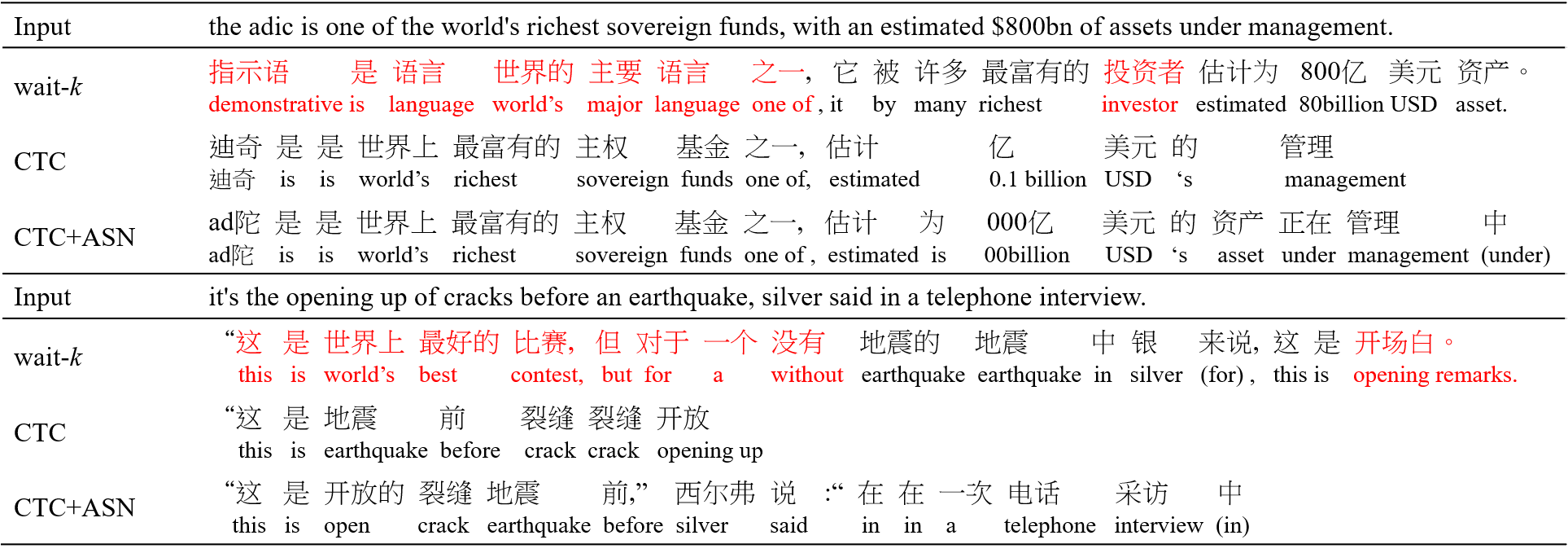}
    \caption{Examples from CWMT En$\to$Zh. Text in red are hallucinations unrelated to source. We use $k=3$ models.}
    \label{fig:examples}
\end{figure*}

We show some examples from the CWMT test set. We compare the predictions from wait-$k$, CTC, and CTC+ASN models in Figure~\ref{fig:examples}. 
In the first example, wait-$k$ predicts the sentence \textit{``demonstrative is one of the major languages in the world's languages,''} which is clearly hallucination. CTC failed to translate \textit{``8000''} and \textit{``assets,''} which shows that CTC may under-translate and ignore source information.  
In the second example, wait-$k$ hallucinates the sentence \textit{``this is the world's best contest, but to a earthquake without earthquake, it's the opening remarks.''} CTC under-translates \textit{``silver said in a telephone interview.''}
Our method generally provides translation that preserves the content. Although our model prediction is a bit less fluent than wait-$k$, they are generally comprehensible. See Appendix~\ref{appendix:more_examples} for more examples.

\begin{figure}[h]
    \centering
    \def\figwid{\linewidth}
    \includegraphics[width=\figwid]{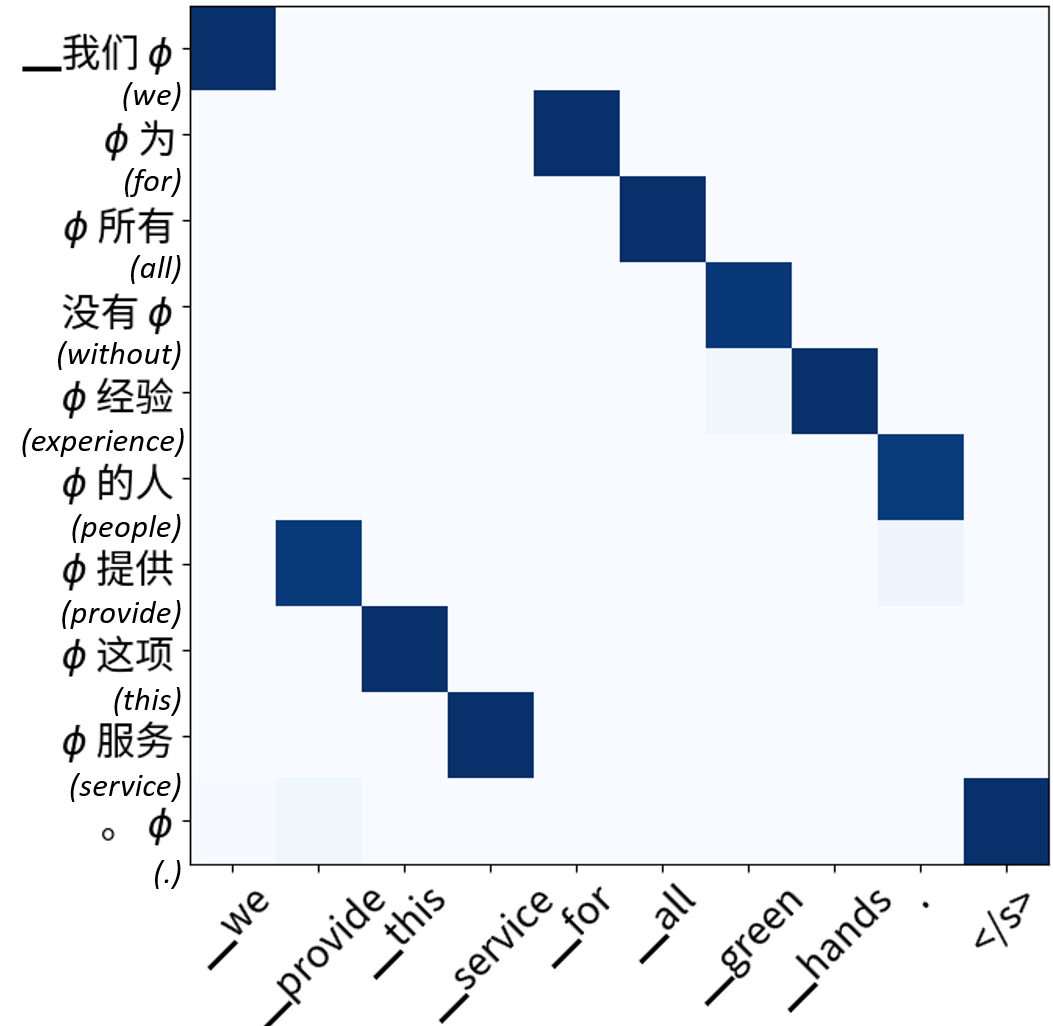}
    \caption{The $\rmz$ predicted by ASN. The horizontal axis is the source tokens. The vertical axis is the output positions, each corresponds to 2 target tokens.}
    \label{fig:permute1}
\end{figure}

We study the output of the ASN to verify that reordering information is being learned. Figure~\ref{fig:permute1} shows an example of the permutation matrix $\rmz$ predicted by the ASN. The horizontal axis is labeled with the source tokens. The vertical axis is the output positions, each are labeled with 2 target tokens (due to the length projection).
In the example, the English phrase \textit{``for all green hands''} come late in the source sentence, but their corresponding Chinese tokens appear early in target, which causes anticipation. 
Our ASN permutes the hidden states of this phrase to early positions, so anticipation no longer happens, and provides the correct training signal for the model. We provide additional examples in Appendix~\ref{appendix:asn_out}. 

%% file: sections/ablation.tex
\section{Ablation Study}\label{sec:ablation}
We perform ablation studies on the CWMT dataset.
\subsection{Gumbel-Sinkhorn Network}\label{sec:ablation-gumbel}
We show that the Gumbel-Sinkhorn Network is crucial to our method. We train CTC+ASN models with $k=3$ under the following settings:\footnote{we do not use weight initialization in this subsection.}
\begin{itemize}
    \setlength\itemsep{0em}
    \item \textbf{No temperature}: Set the temperature $\tau$ to 1.
    \item \textbf{No noise}: Set the Gumbel noise factor $\delta$ to 0.
    \item \textbf{Gumbel softmax}: Replace Sinkhorn normalization with softmax.
    \item \textbf{Default}: The Gumbel-Sinkhorn Network.
\end{itemize}
Table~\ref{tab:ablation_sinkhorn} shows the result of these settings. Without low temperature, the ASN output $\rmz$ is not sparse, which means the content of individual vectors in $\rmh$ is not preserved after applying ASN. Because ASN is removed during inference, this creates a train-test mismatch for the projection network, which is detrimental to the prediction quality ((a) v.s. (d)). Removing the noise ignores the sampling process, which hurts the robustness of the model ((b) v.s. (d)). Using softmax instead of Sinkhorn normalization makes $\rmz$ not doubly stochastic, which means $\overline{\rmh}$ might not cover every vector in $\rmh$. Those not covered are not optimized for generation during training.
However, during inference, all vectors in $\rmh$ are passed to length projection to generate tokens. This mismatch is also harmful to the result ((c) v.s. (d)).

\begin{table}[h]
\centering
\begin{tabular}{@{}lc@{}}
\toprule
\multicolumn{1}{c}{\textbf{Settings}} & \textbf{BLEU($\uparrow$)} \\ \midrule
(a) No temperature & 28.39 \\
(b) No noise & 27.88 \\
(c) Gumbel softmax & 36.54 \\
(d) Default & \textbf{38.92} \\ \bottomrule
\end{tabular}
\caption{Test set BLEU scores of different settings. }
\label{tab:ablation_sinkhorn}
\end{table}

\subsection{Weight Initialization}
\label{sec:ablation-weight-init}

\begin{figure}[h]
    \centering
    \def\figwid{\linewidth}
    \includegraphics[width=\figwid]{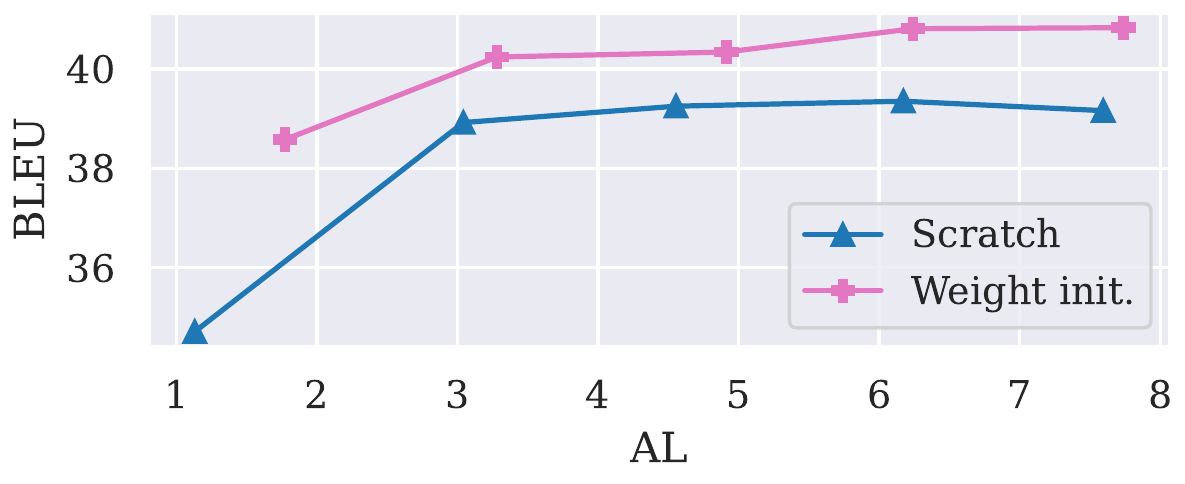}
    \caption{Latency and quality comparison between the model trained from scratch and one with weight initialization.}
    \label{fig:ablation_init}
\end{figure}

\begin{figure}[h]
    \centering
    \def\figwid{0.9\linewidth}
    \includegraphics[width=\figwid]{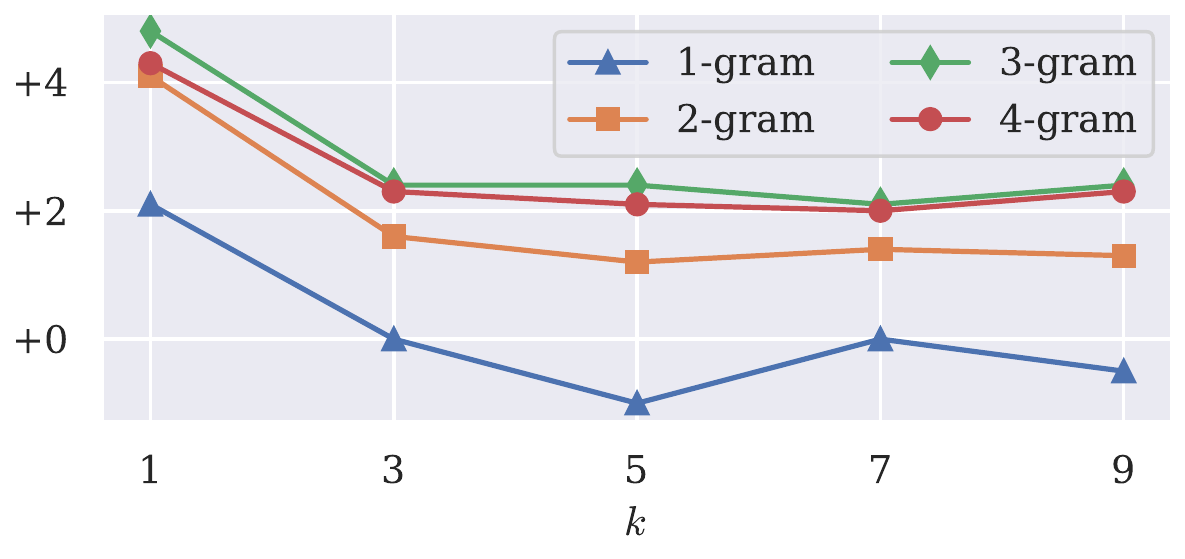}
    \caption{The n-gram precision improvement of weight initialization compared to Scratch across different delays ($k$).}
    \label{fig:ngram}
\end{figure}

We investigate the effectiveness of initializing encoder parameters from the CTC baseline model. Specifically, we train the CTC+ASN model from scratch to compare it with the weight initialized setting. 
As Figure~\ref{fig:ablation_init}
reveals, the weight initialization significantly improves the translation quality while slightly increasing the latency. 
\par
This improvement comes from what was already learned by the CTC baseline model. The CTC baseline model learns to perform reordering, i.e., it outputs blank symbols when reading the information, then outputs the content in the target language order. Such information might span several source tokens,
so the AL of the CTC baseline model is high (Figure~\ref{fig:latency-quality-enzh}).
In our weight initialized setting, ASN handles the long-distance reordering that CTC was struggling with, while the local reordering already learned by CTC is preserved. 
In contrast, when trained from scratch, ASN would learn most of the reordering, so the encoder would not learn to perform local reordering. 
We hypothesize that if the model performs local reordering during inference, its latency might increase, but the higher order n-grams precision can improve, which benefits its quality.
Indeed, Figure~\ref{fig:ngram} indicates that the weight initialization mostly improves the 2,3,4-gram precision of the BLEU score. 

%% file: sections/conclusion.tex
\section{Conclusion}
We proposed a framework to alleviate the impact of long-distance reordering on simultaneous translation. We apply our method to the CTC model and show that it improves the translation quality and latency, especially English to Chinese translation. We verified that the ASN indeed learns the correct alignment between source and target. Besides, we showed that a single encoder can perform simultaneous translation with competitive quality in low latency settings and enjoys the speed advantage over wait-$k$ Transformer.

%% file: sections/appendix.tex
\clearpage
\section{Source Code}
\label{appendix:code}
Our source code is available at \url{https://github.com/George0828Zhang/sinkhorn-simultrans}. Please follow the instructions in \texttt{README.md} to reproduce the results.

\section{Datasets}
\label{appendix:datasets}
We use the CWMT English to Chinese and WMT15 German to English datasets for experiments. They can be downloaded in the following links: 1) CWMT \url{http://nlp.nju.edu.cn/cwmt-wmt/}) 2) WMT15 \url{http://www.statmt.org/wmt15/translation-task.html}. The WMT15 De-En is a widely used corpus for simultaneous machine translation, in the news domain. Another popular dataset is the NIST En-Zh corpus, however, NIST is not publicly available, thus we use CWMT corpus instead. CWMT is also in the news domain.
\par
Both datasets are publicly available. We didn't find any license information for both. We adhered to the terms of use for both. We didn't find any information on names or uniquely identified individual people or offensive content and the steps taken to protect or anonymize them.

\section{Transformer Hyperparameters}
\label{appendix:hyper}
Our architecture related hyperparameters are listed in Table~\ref{tab:hyperparams_arch}. We follow the \textit{base} configuration of Transformer for encoder-decoder models. For models without decoder, we follow the same configuration for its encoder. 
The total parameter count for Transformer is 76.9M. For encoder-only models without ASN, it is 52.2M. The ASN has 12.6M parameters.


\begin{table}[h!]
\centering
\begin{tabular}{@{}cccc@{}}
\toprule
Hyperparameter   & (A)         & (B) \\ \midrule
encoder layers   & 6           & 6         \\
decoder layers   & 6           & 0         \\
embed dim        & 512         & 512       \\
feed forward dim & 2048        & 2048      \\
num heads        & 8           & 8         \\
dropout          & 0.1         & 0.1       \\ \bottomrule
\end{tabular}
\caption{Transformer architecture related hyperparameters for each model. (A) full-sentence and wait-$k$ model (B) CTC encoder model.}
\label{tab:hyperparams_arch}
\end{table}

\section{ASN Hyperparameters}
\label{appendix:asn_hyper}
We perform a Bayesian hyperparameter optimization on both datasets using the sweep utility provided by Weights \& Biases~\cite{wandb}. Table~\ref{tab:hyperparams_asn} shows the search range and the selected values. We found a well performing set in the 7th run for CWMT and 1st run for WMT15. It is possible that different k might prefer different hyperparameters. However, we use the same set to
fairly compare to wait-k, and to reduce the cost.
All subsequent results are obtained using this set of values if not specified.

\begin{table}[h]
\centering
\adjustbox{max width=\linewidth}{%
\begin{tabular}{@{}cccc@{}}
\toprule
Hyperparameter            & CWMT  & WMT15 & Range   \\ \midrule
layers $M$     & 3     & 3 & 1, 3                   \\
iterations $l$            & 16    & 16&  4, 8, 16                  \\
temperature $\tau$        & 0.25   & 0.13 & {[}0.05, 0.3{]} \\
noise factor $\delta$     & 0.3   & 0.45&{[}0.1, 0.3{]} \\
upsample ratio $\mu$      & 2     & 2& 2, 3                     \\ 
mask ratio $\gamma$       & 0.5   & 0.5& {[}0., 0.7{]} \\ \bottomrule
\end{tabular}}
\caption{ASN related hyperparameters and the search range. We use Bayesian hyperparameter optimization, so the combinations are not exhaustively searched.}
\label{tab:hyperparams_asn}
\end{table}

\section{Hardware and Environment}
\label{appendix:hardware}
For training, each run are conducted on a container with a single Tesla V100-SXM2-32GB GPU, 4 CPU cores and 90GB memory. The operating system is \url{Linux-3.10.0-1127.el7.x86_64-x86_64-with-glibc2.10}. The version of Python is 3.8.10, and version of PyTorch is 1.9.0. We use a specific version of fairseq~\cite{ott-etal-2019-fairseq} toolkit, the instructions are provided in \texttt{README.md} of our source code. All run uses mixed precision (i.e. fp16) training implemented by fairseq. All training took 10-15 hours to converge (early stopped).
\par
For inference, the evaluation are conducted on another machine with 12 CPU cores (although we restrict the evaluation to only use 2 threads), 32GB memory and no GPU is used. The operating system is \url{Linux-5.11.0-25-generic-x86_64-with-glibc2.10}.

\section{Gumbel-Sinkhorn Operator}
\label{appendix:gumbel_sinkhorn}
The Sinkhorn normalization~\cite{adams2011ranking} iteratively performs row-wise and column-wise normalization on a matrix, converting it to a doubly stochastic matrix.
Formally, for a $N$ dimensional square matrix $X\in\mathbb{R}^{N\times N}$, the Sinkhorn normalization $S(X)$ is defined as:
\begin{align}
    S^{0}(X) &= \exp(X), \\
    S^{l}(X) &= \mathcal{T}_c\left(
                    \mathcal{T}_r\left( S^{l-1}(X) \right)
                \right), \\
    S(X) &= \lim_{l\to\infty} S^{l}(X). \label{eq:converge_dsm}
\end{align}
where $\mathcal{T}_r$ and $\mathcal{T}_c$ are row-wise and column-wise normalization operators on a matrix, defined below:
\begin{align}
    \mathcal{T}_r(X) &= X \oslash (X\colvecone_{N}\colvecone_{N}^{\top}), \\
    \mathcal{T}_c(X) &= X \oslash (\colvecone_{N}\colvecone_{N}^{\top}X).
\end{align}
The $\oslash$ denotes the element-wise division, and $\colvecone_{N}$ denotes a column vector full of ones. 
As the number of iterations $l$ grows, $S^{l}(X)$ will eventually converge to a doubly stochastic matrix (equation~\ref{eq:converge_dsm})~\cite{sinkhorn1964relationship}. In practice, we often consider the truncated version, where $l$ is finite.
\par
On the other hand, the Gumbel-Sinkhorn operator adds the Gumbel reparametrization trick~\cite{kingma2013auto} to the Sinkhorn normalization, in order to approximate the sampling process. 
It can be used to estimate marginal probability via sampling.
Formally, suppose that a noise matrix $\varepsilon$ is sampled from independent and identically distributed (i.i.d.) Gumbel distributions:
\begin{equation}
    \rmeps\in\mathbb{R}^{N\times N}  \overset{i.i.d.}{\sim} Gumbel(0, 1).
\end{equation}
The Gumbel-Sinkhorn operator is described by first adding the Gumbel noise $\rmeps$, then scaling by a positive temperature $\tau$, and finally applying the Sinkhorn normalization:
\begin{equation}
    S((X+\rmeps)/\tau).\label{eq:gumbel_sinkhorn_op}
\end{equation}
By taking the limit $\tau\to0^+$, the output converges to a permutation matrix. The Gumbel-Sinkhorn operator approximates sampling from a distribution of permutation matrices. Thus, the equation~\ref{eq:marginalize_conditional} can be estimated through sampling:
\begin{equation}
    p(\rmy|\rmx) = \mathbb{E}_{\rmz\sim p(\rmz|\rmx)} \left[p_g(\rmy|\rmx, \rmz)\right].
\end{equation}
In practice, we sample from $p(\rmz|\rmx, \rmy)$ instead, as it is easier to perform word alignment ($p(\rmz|\rmx, \rmy)$) than directly predicting order ($p(\rmz|\rmx)$).

\section{Details on Evaluation Metrics}
\label{appendix:detail_metrics}
\subsection{Average Lagging (AL)}
The AL measures the degree the user is out of sync with the speaker~\cite{ma2019stacl}. It measures the system's lagging behind an oracle wait-0 policy. For a read-write policy $g(\cdot)$, define the cut-off step $\tau_g(|\rmx|)$ as the decoding step when source sentence finishes: 
$$\tau_g(|\rmx|)=\min\{t|\quad g(t)=|\rmx|\}$$
Then the AL for an example $\rmx, \rmy$ is defined as:
$$
\mathrm{AL}_g(\rmx, \rmy)=\frac{1}{\tau_g(|\rmx|)}\sum_{t=1}^{\tau_g(|\rmx|)} g(t) - \frac{t-1}{|\rmy|/|\rmx|}
$$
The second term in the summation represents the ideal latency of an oracle wait-0 policy in terms of target words (or characters for Chinese). The AL averaged across the test set is reported.

\subsection{Computation Aware Average Lagging (AL-CA)}
Originally proposed for simultaneous speech-to-text translation~\cite{ma2020simulmt}, the AL-CA is similar to AL, but takes the actual computation time into account, and is measured in milliseconds.
\begin{align}
&\mathrm{AL}_g^{CA}(\rmx, \rmy) \nonumber\\
&=\frac{1}{\tau_g(|\rmx|)}\sum_{i=1}^{\tau_g(|\rmx|)} d_{CA}(y_i) - \frac{(i-1)\cdot T_s}{|\rmy|/|\rmx|}
\end{align}
The $d_{CA}(y_i)$ is the the time that elapses from the beginning of the process to the prediction of $y_i$, \textbf{which considers computation}. $T_s$ represents the actual duration of each source feature. The second term in the summation represents the ideal latency of an oracle wait-0 policy in terms of milliseconds, \textbf{without considering computation}. In speech-to-text translation, $T_s$ corresponds to the duration of each speech feature. However, since our source feature is text, the ``actual duration'' for a word is unavailable, so we set $T_s=1$. 
\par
The motivation behind using AL-CA here is to show the speed advantage of CTC models. When calculating AL-CA, we account for variance by running the evaluation 3 times and report the average.

\subsection{Character n-gram F-score (chrF)}
The general formula for the chrF score is given by:
\begin{equation}
    \mathrm{chrF}\beta = (1+\beta^2) \frac{\mathrm{chrP}\cdot \mathrm{chrR}}{\beta^2\cdot \mathrm{chrP}+\mathrm{chrR}}.
\end{equation}
where
\begin{itemize}
    \item chrP: percentage of character n-grams in the hypothesis which have a counterpart in the reference.
    \item chrR: percentage of character n-grams in the reference which are also present in the hypothesis.
    \item $\beta$: a parameter which assigns $\beta$ times more importance to recall than to precision.
\end{itemize}
The maximum n-gram length $N$ is optimal when $N=6$~\cite{popovic2015chrf}, and the optimal $\beta$ is shown to be $\beta=2$~\cite{popovic2016chrf}. 
\par
The motivation behind using chrF2 is that 1) as machine translation researchers, we are encouraged to report multiple automatic evaluation metrics. 2) BLEU is purely precision-based, while chrF2 is F-score based, which takes recall into account. 3) chrF2 is shown to correlate better with human rankings than the BLEU score.

\subsection{$k$-Anticipation Rate ($k$-AR)}
\begin{figure}[h]
    \centering
    \def\figwid{0.45\textwidth}
    \includegraphics[width=\figwid]{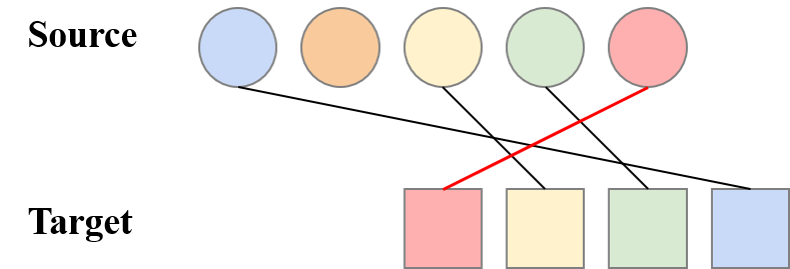}
    \caption{An example of 2-anticipation. The links are alignments, and the red link is an instance of anticipation.}
    \label{fig:kar-explained}
\end{figure}

For each sentence pair, we first use \textit{awesome-align}~\cite{dou2021word} to extract word alignments, then for each aligned target word $y_j$, it is considered a $k$-anticipation if it is aligned to a source word $x_i$ that is $k$ words behind, in other words, if $i-k+1>j$. See Figure~\ref{fig:kar-explained} for an example of 2-anticipation. The $k$-AR is calculated as the percentage of $k$-anticipation among all aligned word pairs.

\section{SimulEval Configuration}
\label{appendix:simuleval_config}
Table~\ref{tab:simuleval_config} show the language specific options for latency evaluation on SimulEval, which affect the AL calculation.

\begin{table}[h]
\centering
\begin{tabular}{@{}ccc@{}}
\toprule
\textbf{Options} & \textbf{En} & \textbf{Zh} \\ \midrule
--eval-latency-unit & word & char \\ \midrule
--no-space & false & true \\ \bottomrule
\end{tabular}
\caption{Configuration for SimulEval under different target languages.}
\label{tab:simuleval_config}
\end{table}

\section{SacreBLEU Signatures}
\label{appendix:signature}
Table~\ref{tab:signature} shows the signatures of SacreBLEU evaluation.
\begin{table}[h]
\centering
\begin{tabular}{ccc}
\hline
Lang  & Metric & Signature                                                                                                                                \\ \hline
Zh    & BLEU   & \begin{tabular}[c]{@{}c@{}}nrefs:var\textbar bs:1000\textbar seed:12345\\ \textbar case:lc\textbar eff:no\textbar tok:zh\\ \textbar smooth:exp\textbar version:2.0.0\end{tabular}                \\ \hline
Zh    & chrF2    & \begin{tabular}[c]{@{}c@{}}nrefs:var\textbar bs:1000\textbar seed:12345\\ \textbar case:lc\textbar eff:yes\textbar nc:6 \textbar nw:0 \\ \textbar space:no\textbar version:2.0.0\end{tabular} \\ \hline
En & BLEU   & \begin{tabular}[c]{@{}c@{}}nrefs:1\textbar bs:1000\textbar seed:12345\\ \textbar case:lc\textbar eff:no\textbar tok:13a\\ \textbar smooth:exp\textbar version:2.0.0\end{tabular}                \\ \hline
En & chrF2    & \begin{tabular}[c]{@{}c@{}}nrefs:1\textbar bs:1000\textbar seed:12345\\ \textbar case:lc\textbar eff:yes\textbar nc:6 \textbar nw:0 \\ \textbar space:no\textbar version:2.0.0\end{tabular} \\ \hline
\end{tabular}
\caption{The SacreBLEU signatures for each target language and each metric.}
\label{tab:signature}
\end{table}

\section{Detailed Statistics of Quality Metrics}
\label{appendix:detailed_results}
Table~\ref{tab:detailed_results} shows the detailed distributional statistics of the quality metrics evaluated on the CWMT and WMT15 datasets. All settings are trained once, but we use statistical significant test using bootstrap resampling.

\section{Latency-quality results with chrF}
\label{appendix:chrf_results}
\begin{figure*}[h]
    \centering
    \def\figwid{0.49\textwidth}
    \includegraphics[width=\figwid]{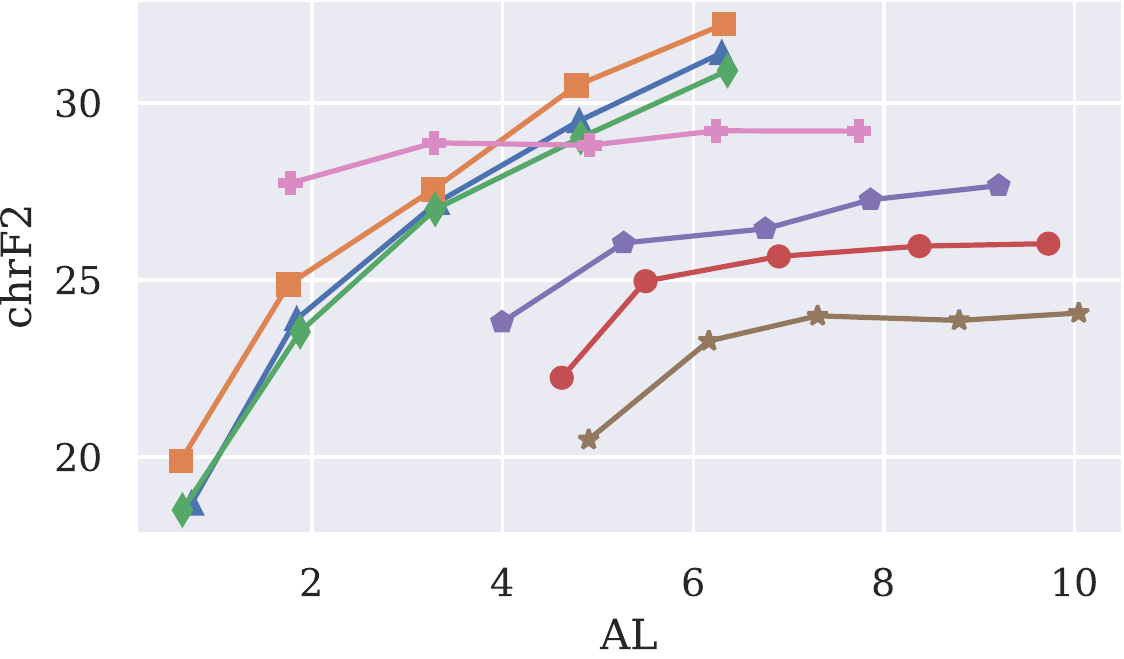}
    \includegraphics[width=\figwid]{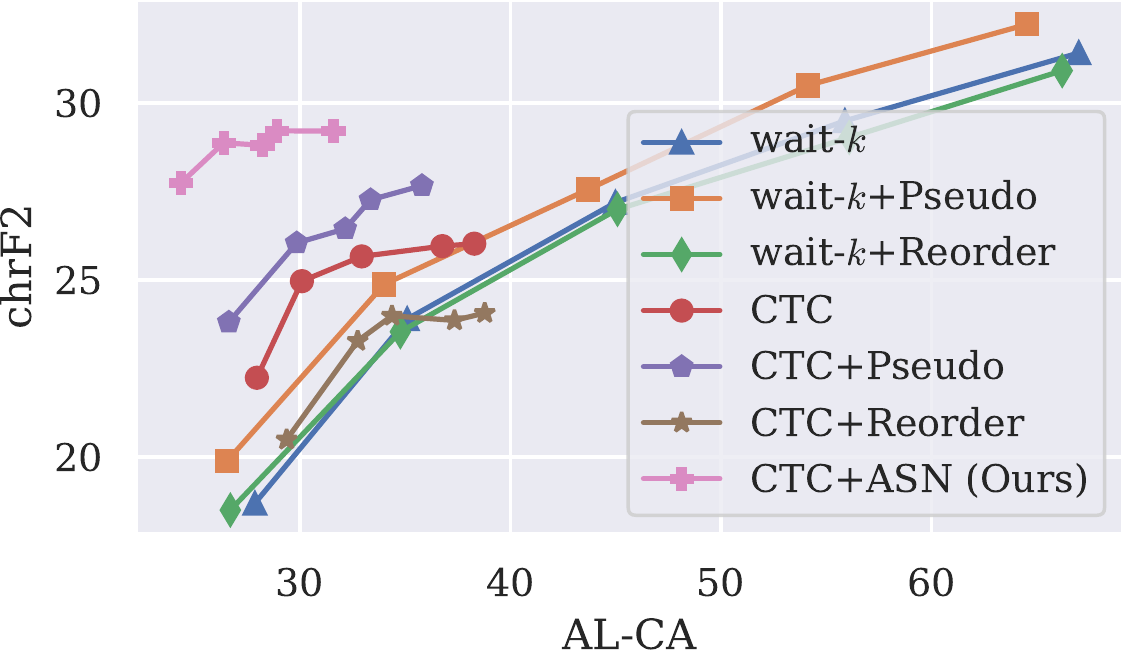}
    
    \caption{Latency-quality trade off with chrF score on the CWMT En-Zh dataset. Each line represents a system, and the 5 nodes corresponds to $k=1,3,5,7,9$, from left to right. The figures share the same legend.}
    \label{fig:latency-quality-chrf-enzh}
\end{figure*}

\begin{figure*}[h]
    \centering
    \def\figwid{0.49\textwidth}
    \includegraphics[width=\figwid]{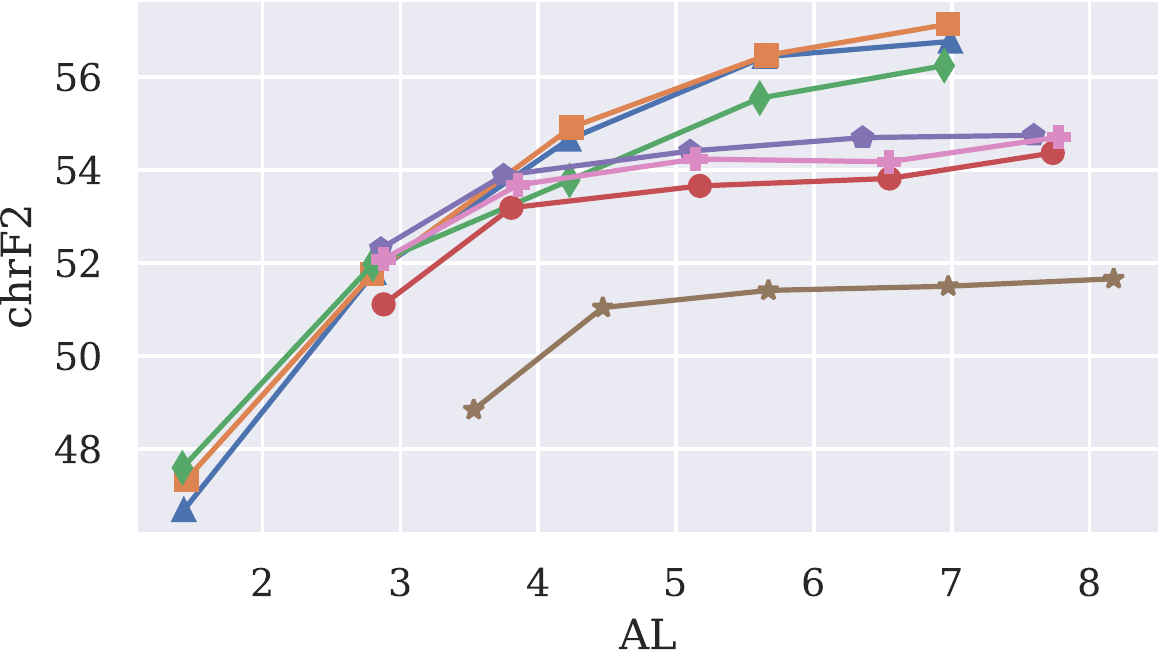}
    \includegraphics[width=\figwid]{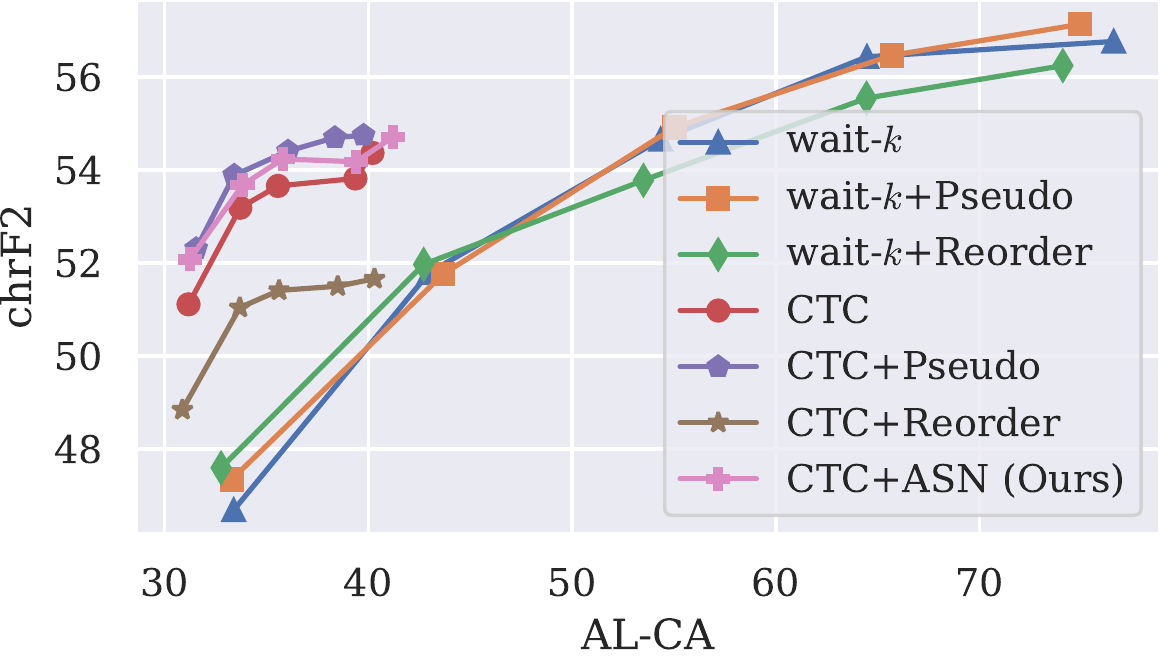}
    
    \caption{Latency-quality trade off with chrF score on the WMT15 De-En dataset. Each line represents a system, and the 5 nodes corresponds to $k=1,3,5,7,9$, from left to right. The figures share the same legend.}
    \label{fig:latency-quality-chrf-deen}
\end{figure*}

Figure~\ref{fig:latency-quality-chrf-enzh} show the quality-latency trade off with chrF on the CWMT En-zh dataset.
Figure~\ref{fig:latency-quality-chrf-deen} show the quality-latency trade off with chrF on the WMT15 De-En dataset. These results have similar trends with BLEU score.


\section{Performance with Oracle Reordering}
\label{appendix:oracle}
We study our encoder models' performance when the oracle reordering is provided. To achieve this, we re-use the ASN during inference, and fed the (first) reference translation as the context to ASN to estimate $\rmz$. The results compared to default setting is shown in Table~\ref{tab:results-oracle}. This result serves as a upperbound for the performance of CTC-based encoder models.

\begin{table}[t]
\centering
\adjustbox{max width=\linewidth}{%
\begin{tabular}{@{}clccc@{}}
\toprule
\textbf{$k$} & \multicolumn{1}{c}{\textbf{Method}} & \textbf{BLEU} & \textbf{1/2/3/4-gram} & \textbf{BP} \\ \midrule
\multirow{2}{*}{$1$}          & Default  & 38.58 & 76.7 / 51.0 / 32.5 / 20.6 & 0.96 \\
                                & + Oracle & 41.59 & 76.0 / 52.7 / 35.9 / 23.9 & 0.96 \\ \midrule
\multirow{2}{*}{$3$}          & Default  & 40.24 & 79.5 / 53.7 / 34.8 / 22.6 & 0.94 \\
                                & + Oracle & 41.75 & 77.5 / 53.7 / 36.5 / 24.4 & 0.95 \\ \midrule
\multirow{2}{*}{$5$}          & Default  & 40.34 & 78.8 / 53.5 / 35.0 / 22.7 & 0.94 \\
                                & + Oracle & 41.70 & 76.0 / 52.4 / 35.5 / 23.6 & 0.98 \\ \midrule
\multirow{2}{*}{$7$}          & Default  & 40.81 & 80.0 / 54.2 / 35.2 / 22.9 & 0.94 \\
                                & + Oracle & 43.37 & 78.8 / 55.2 / 37.9 / 25.8 & 0.96 \\ \midrule
\multirow{2}{*}{$9$}            & Default  & 40.83 & 79.5 / 54.1 / 35.4 / 23.1 & 0.94 \\
                                & + Oracle & 41.77 & 76.3 / 52.7 / 35.5 / 23.6 & 0.98 \\ \bottomrule
\end{tabular}
}
\caption{The BLEU score on the CWMT dataset, including n-gram precision and brevity penalty (BP), of the CTC+ASN system for each $k$ with and without oracle order.}
\label{tab:results-oracle}
\end{table}


\section{More on ASN Output}
\label{appendix:asn_out}
We describe how the target tokens are placed on the vertical axis of the ASN output illustration. Since the length projection upsamples $\overline{\rmh}$ to 2 times longer, each position of $\overline{\rmh}$ corresponds to two target tokens (including repetition and blank symbols introduced by CTC). To find the optimal position for each target tokens and blank symbols, we use the Viterbi alignment (an implementation is publicly available at \url{https://github.com/rosinality/imputer-pytorch}) to align the model's logits and the actual target tokens.

\begin{figure*}[h]
    \centering
    \subcaptionbox{\label{fig:permute2}}{\includegraphics[width=0.48\textwidth]{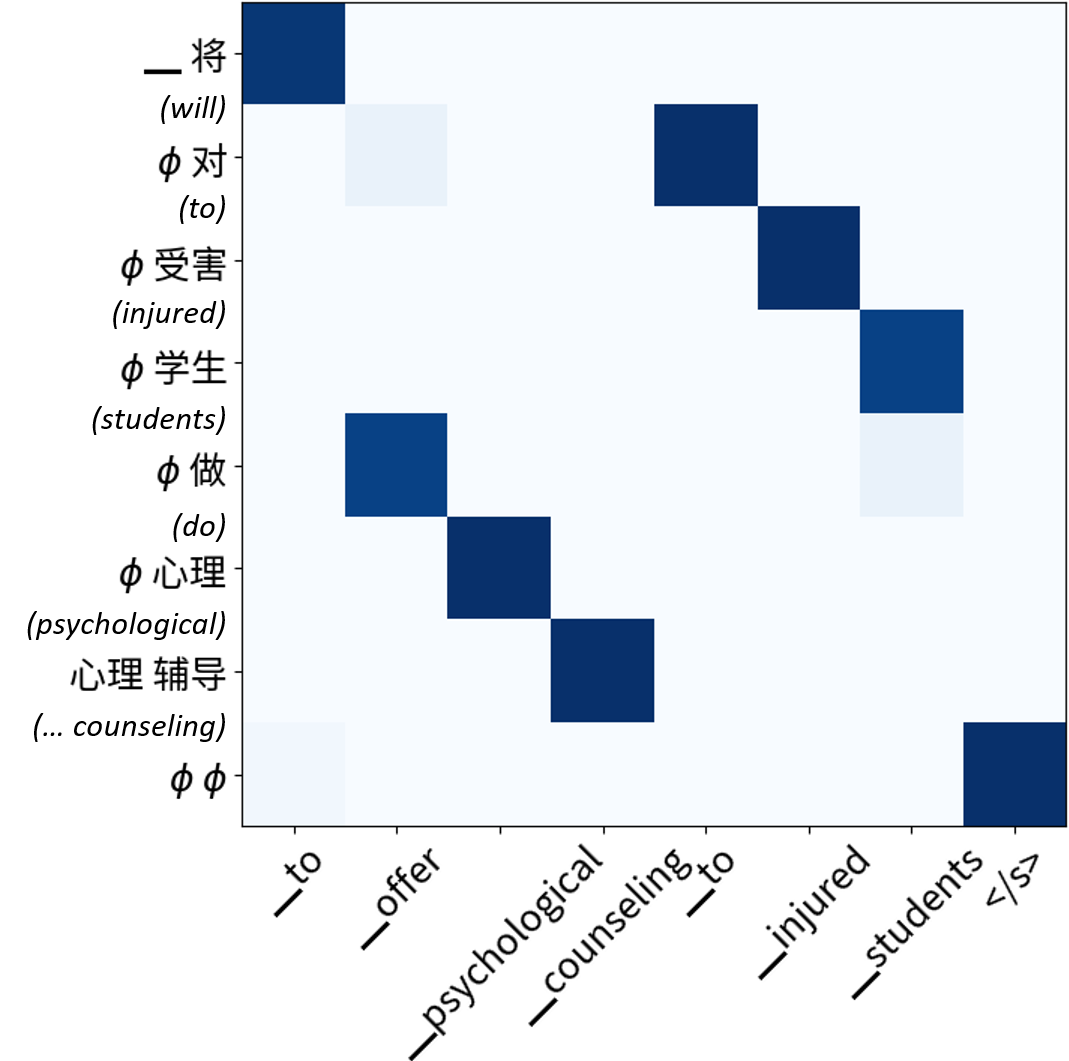}}
    \subcaptionbox{\label{fig:permute3}}{\includegraphics[width=0.5\textwidth]{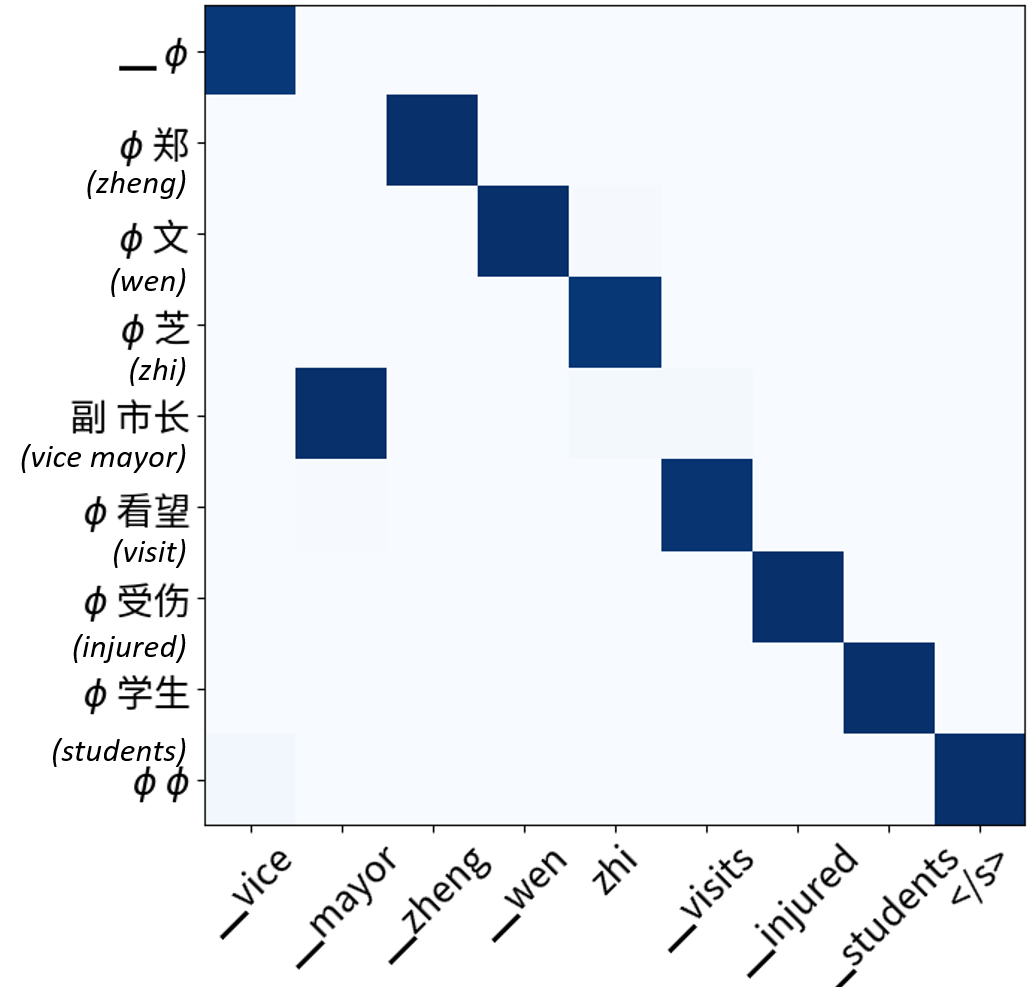}}\vfill
    \subcaptionbox{\label{fig:permute4}}{\includegraphics[width=0.52\textwidth]{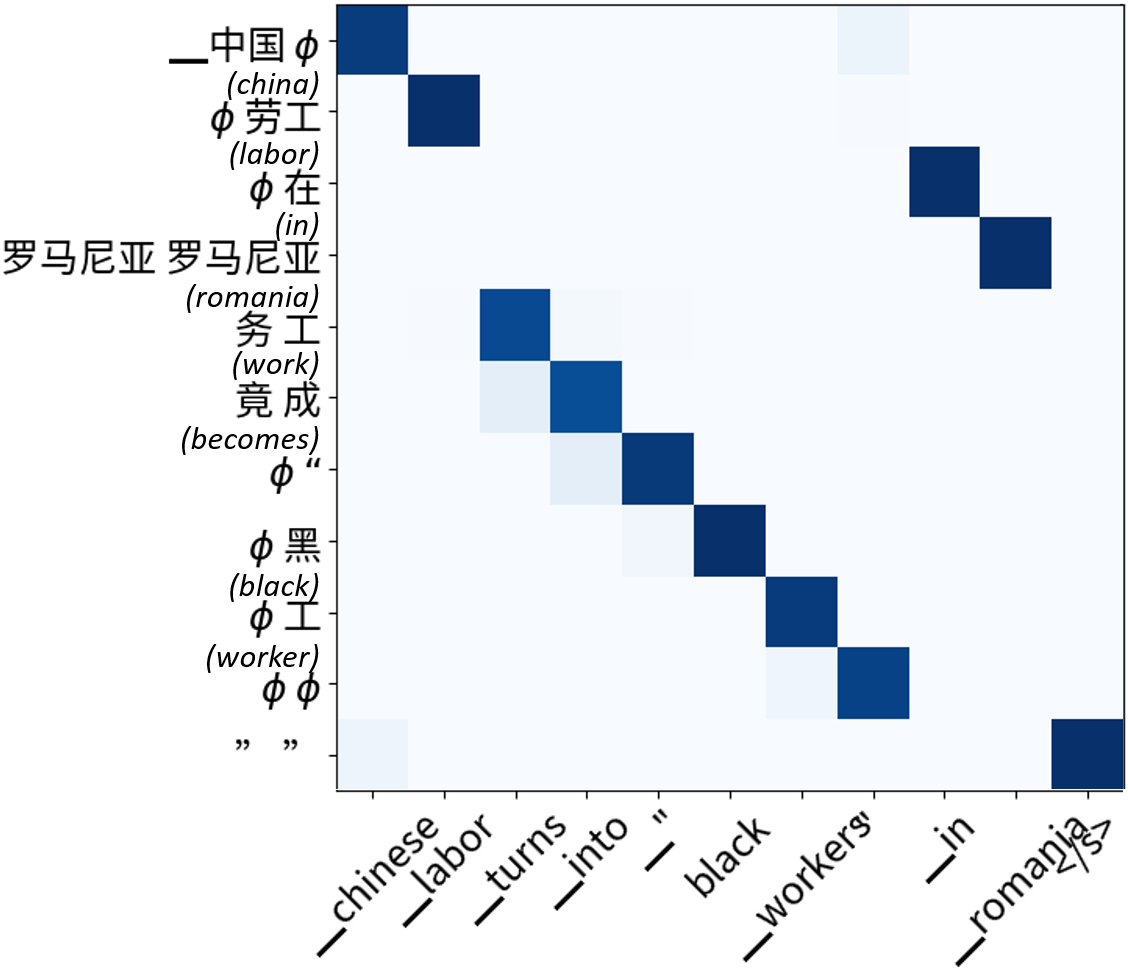}}
    \subcaptionbox{\label{fig:permute5}}{\includegraphics[width=0.46\textwidth]{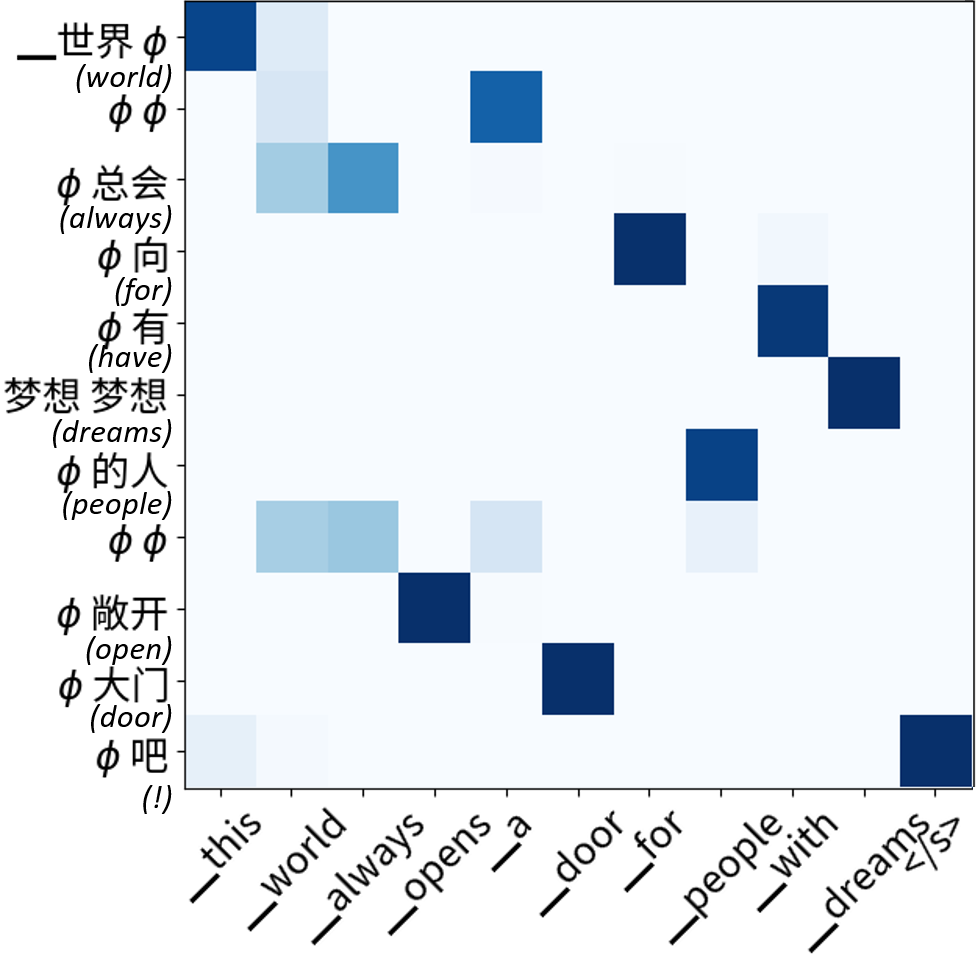}}
    \caption{More approximated permutation matrices predicted by ASN.}
    \label{fig:more_permute}
\end{figure*}

Figure~\ref{fig:more_permute} shows more examples of the approximated permutation matrix predicted by the ASN. The sentence pairs are from CWMT En-Zh test set.


\section{More CWMT Examples}
\label{appendix:more_examples}
\begin{figure*}[h]
    \centering
    \def\figwid{\textwidth}
    \includegraphics[width=\figwid]{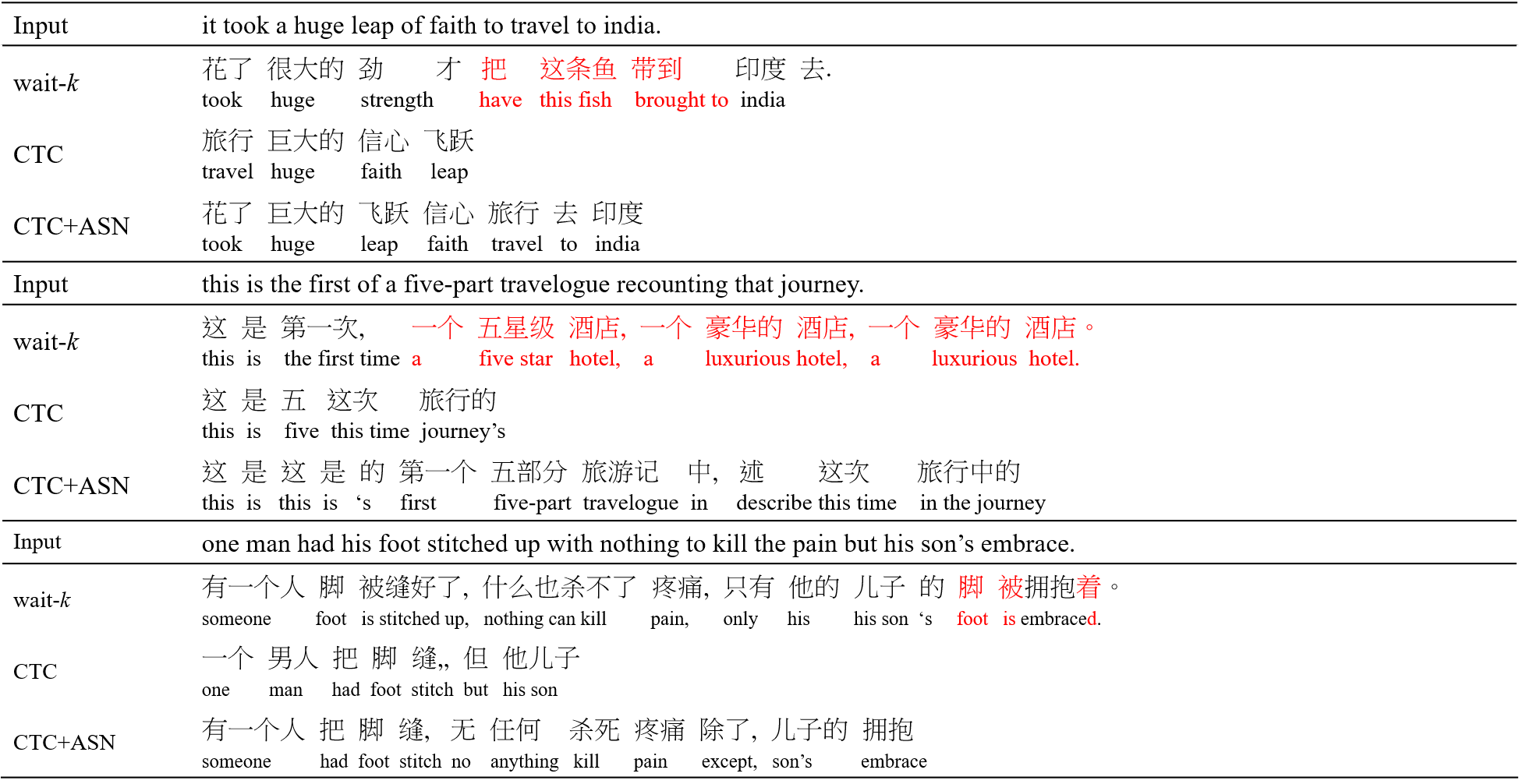}
    \includegraphics[width=\figwid]{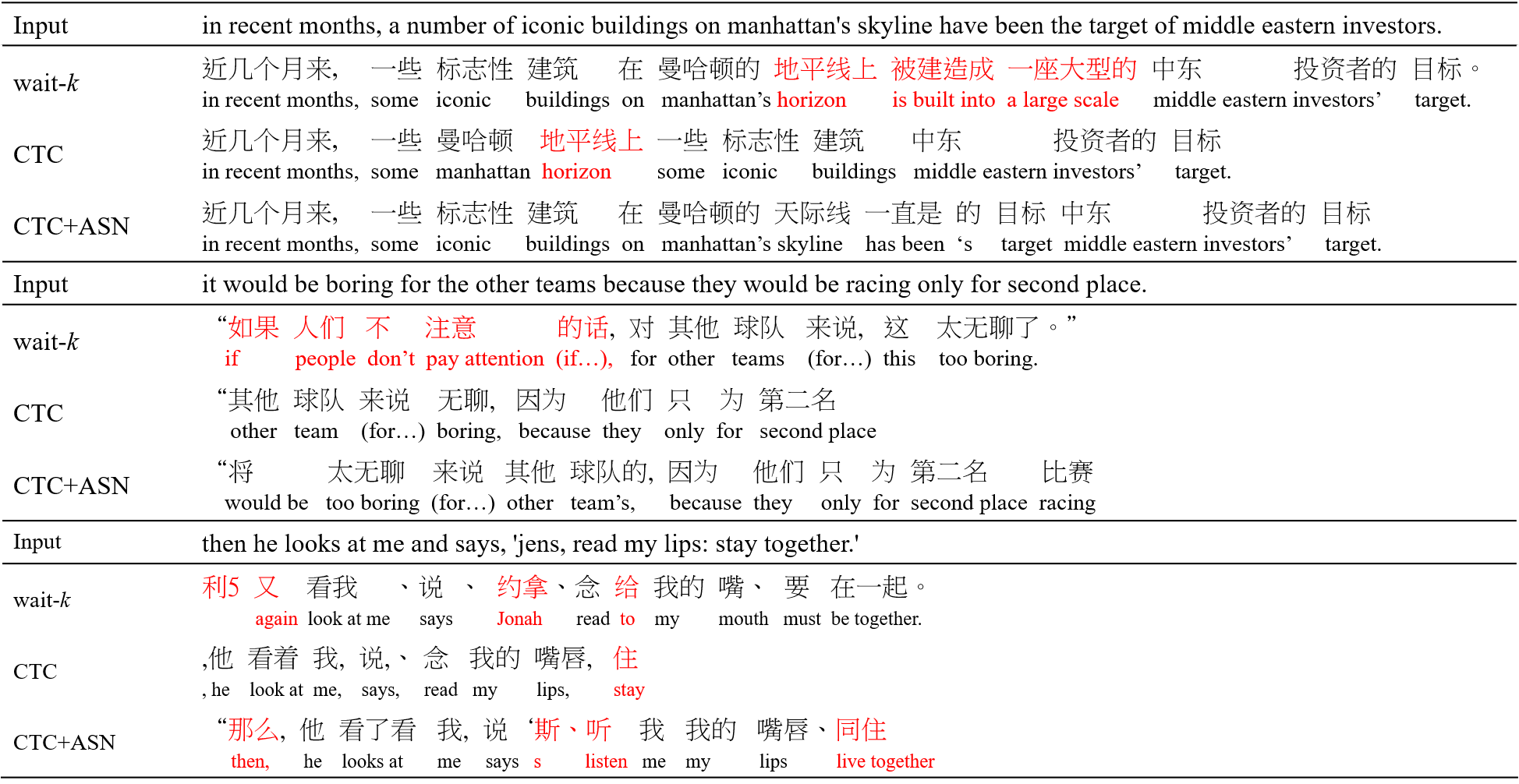}
    \caption{More examples from CWMT En$\to$Zh. Text in red are hallucinations unrelated to source. We use $k=3$ models.}
    \label{fig:more_examples}
\end{figure*}

Figure~\ref{fig:more_examples} shows more examples from CWMT test set and the predictions of wait-$k$, CTC and CTC+ASN models.

\section{FAQ}
\label{appendix:faq}
\renewcommand{\thesubsection}{Q\arabic{subsection}}
\subsection{The trained ASN cannot be used during inference, how to guarantee the model can still perform reordering?}
We categorize reordering into local reordering and long-distance reordering. Our goal is for the ASN to primarily deal with long-distance reordering. In Section~\ref{sec:ablation-weight-init}, we observed that employing the weight initialization improves the 2,3,4-gram precision (but not the unigram), and slightly increases the latency. This suggest that CTC+ASN model can indeed perform local reordering during inference.
\par
As for long-distance reordering, we stress that in simultaneous interpretation, humans actively avoid long-distance reordering in order to reduce latency, which is also the goal of SimulMT. This provides the justification for removing the ASN during inference. (equation~\ref{eq:remove_z})
\par
We additionally provide the performance when $\rmz$ is available during inference in Appendix~\ref{appendix:oracle}.

\subsection{Using ASN during training may cause the model to rely on $\rmz$, which may cause train-test discrepancy during inference?}
In terms of the mismatch of hidden representation, because Gumbel-Sinkhorn gaurantees that $\rmz$ is doubly stochastic (and almost permutation, depending on $\tau$), the representation before and after ASN would only differ by a permutation. This is also discussed in Section~\ref{sec:ablation-gumbel} where removing Sinkhorn nomalization indeed negatively impact the performance.
\par
As for the mismatch of the order of the representation, we note that the length projection network is merely a position-wise affine transformation, which means it is independent of time, so the mismatch of order between training and testing would not negatively impact the prediction made by the length projection network.

\subsection{Proposed method underperform wait-$k$ in high latency.}\label{faq:high-latency}
Simultaneous translation aims to translate in a short time, hence our work focuses on improving the translation quality under low latency setting. 
The higher latency model is less acceptable in practice. For instance, a $k=9$ model decodes a single word after seeing 9 words. We included the results for experimental completeness purpose.

For the reason why proposed method underperform wait-$k$ model: Based on the observation in Appendix~\ref{appendix:oracle}, 43.37 is the best performance of CTC+ASN method. It is inferior to the wait-9 model's 43.80. We suspect that it is caused by the inherent difference between non-autoregressive (NAR) model and auto-regressive (AR) model. 
However, CTC+ASN method's performance is relatively consistent when the latency decreases, while wait-$k$'s performance decreases drastically. 
Therefore, to fit the simultaneous translation setting, our proposed method is more suitable than wait-k.


\subsection{Explanation for why ASN could outperform Reorder and Pseudo reference baselines?}
For the Reorder baseline, we suspect that since the external aligner is fixed and not jointly optimized, it may produce incorrect alignments, or miss correct ones, producing wrongful training targets. 
\par
As for the Pseudo reference baseline, there are two problems that might limit its effectiveness. For one, the pseudo reference is produced from a full-sentence model while using a wait-k decoding strategy, which is a train-test discrepancy. For another, in order to compensate for the first issue, the original translation is included as a second target for each example. This leads to the infamous multi-modality problem for non-autoregressive models, which might be harmful to our CTC-based encoder.

\subsection{What are the limitations of the proposed method?}
First of all, for SimulMT to be applicable to a conference setting, we assume a streaming ASR is available. However, we did not account for ASR errors in our SimulMT models.

Second, as discussed in Section~\ref{sec:quantitative}, our method is only effective if the language pair includes sufficient long-distance reordering. For instance, when translation from English to Spanish, we there's hardly any reason to employ our method.

Finally, as discussed in \ref{faq:high-latency}, our method is less advantageous when the latency budget is high.

\subsection{What are the risks of the proposed method?}
One risk is that our method may favor low-latency over high precision, which means that erroneous translation may occur, which might twist the meaning of source sentence. However, latency and quality is inherently a trade-off, and erroneous translation could be mitigated by refinement or post-editing techniques.

\begin{table*}[h]
\centering
\adjustbox{max width=\textwidth}{%
\begin{tabular}{@{}clrrrrrrrr@{}}
\toprule
 & \multicolumn{1}{c}{} & \multicolumn{4}{c}{\textbf{CWMT En$\rightarrow$Zh}} & \multicolumn{4}{c}{\textbf{WMT15 De$\rightarrow$En}} \\ \midrule
\textbf{Delay} & \multicolumn{1}{c}{\textbf{Method}} & \multicolumn{1}{c}{\textbf{BLEU}} & \multicolumn{1}{c}{\textbf{$\mu$±95\%CI}} & \multicolumn{1}{c}{\textbf{chrF2}} & \multicolumn{1}{c}{\textbf{$\mu$±95\%CI}} & \multicolumn{1}{c}{\textbf{BLEU}} & \multicolumn{1}{c}{\textbf{$\mu$±95\%CI}} & \multicolumn{1}{c}{\textbf{chrF2}} & \multicolumn{1}{c}{\textbf{$\mu$±95\%CI}} \\ \midrule
offline & Transformer & 45.85 & 45.85±0.60 & 32.46 & 32.46±0.45 & 31.67 & 31.70±0.77 & 57.65 & 57.67±0.61 \\ \midrule
\multirow{7}{*}{$k=1$} & wait-$k$ & 24.31 & 24.29±0.62 & 18.69 & 18.67±0.43 & 19.91 & 19.91±0.68 & 46.68 & 46.70±0.69 \\
 & wait-$k$+Pseudo & *25.93 & 25.91±0.66 & *19.89 & 19.87±0.46 & *20.63 & 20.63±0.68 & *47.34 & 47.35±0.68 \\
 & wait-$k$+Reorder & 23.98 & 23.96±0.59 & 18.50 & 18.49±0.39 & *20.54 & 20.55±0.65 & *47.59 & 47.61±0.68 \\
 & CTC & 28.44 & 28.42±0.56 & 22.24 & 22.24±0.35 & 23.08 & 23.09±0.69 & 51.11 & 51.13±0.56 \\
 & CTC+Pseudo & $^\dagger$30.77 & 30.75±0.61 & $^\dagger$23.81 & 23.81±0.38 & \textbf{$^\dagger$24.48} & 24.49±0.69 & \textbf{$^\dagger$52.31} & 52.32±0.56 \\
 & CTC+Reorder & $^\dagger$24.09 & 24.08±0.58 & $^\dagger$20.49 & 20.48±0.36 & $^\dagger$20.77 & 20.78±0.65 & $^\dagger$48.84 & 48.85±0.56 \\
 & CTC+ASN & \textbf{$^\dagger$38.58} & 38.57±0.45 & \textbf{$^\dagger$27.74} & 27.73±0.32 & $^\dagger$24.17 & 24.19±0.70 & $^\dagger$52.08 & 52.10±0.54 \\ \midrule
\multirow{7}{*}{$k=3$} & wait-$k$ & 32.27 & 32.25±0.65 & 23.90 & 23.90±0.43 & 25.85 & 25.87±0.78 & 51.79 & 51.81±0.67 \\
 & wait-$k$+Pseudo & *33.53 & 33.52±0.64 & *24.88 & 24.87±0.44 & 25.74 & 25.76±0.77 & 51.76 & 51.78±0.66 \\
 & wait-$k$+Reorder & *31.47 & 31.46±0.66 & *23.54 & 23.54±0.45 & *25.26 & 25.28±0.73 & 51.97 & 51.99±0.65 \\
 & CTC & 32.45 & 32.44±0.61 & 24.97 & 24.96±0.39 & 26.07 & 26.09±0.69 & 53.19 & 53.21±0.58 \\
 & CTC+Pseudo & $^\dagger$34.03 & 34.03±0.61 & $^\dagger$26.05 & 26.05±0.39 & \textbf{$^\dagger$26.61} & 26.63±0.68 & \textbf{$^\dagger$53.89} & 53.91±0.55 \\
 & CTC+Reorder & $^\dagger$28.52 & 28.50±0.62 & $^\dagger$23.28 & 23.28±0.40 & $^\dagger$23.50 & 23.52±0.71 & $^\dagger$51.04 & 51.06±0.55 \\
 & CTC+ASN & \textbf{$^\dagger$40.24} & 40.23±0.51 & \textbf{$^\dagger$28.88} & 28.87±0.34 & $^\dagger$26.53 & 26.55±0.73 & $^\dagger$53.68 & 53.70±0.57 \\ \midrule
\multirow{7}{*}{$k=5$} & wait-$k$ & 37.40 & 37.39±0.65 & 27.19 & 27.19±0.44 & \textbf{28.52} & 28.54±0.82 & \textbf{54.66} & 54.68±0.64 \\
 & wait-$k$+Pseudo & *37.96 & 37.95±0.67 & *27.56 & 27.56±0.46 & \textbf{28.68} & 28.71±0.78 & \textbf{54.92} & 54.95±0.60 \\
 & wait-$k$+Reorder & *36.86 & 36.84±0.65 & 27.00 & 26.99±0.44 & *27.35 & 27.38±0.75 & *53.78 & 53.81±0.63 \\
 & CTC & 33.64 & 33.63±0.62 & 25.67 & 25.66±0.39 & 26.51 & 26.53±0.77 & 53.66 & 53.68±0.58 \\
 & CTC+Pseudo & $^\dagger$34.65 & 34.64±0.61 & $^\dagger$26.45 & 26.45±0.40 & $^\dagger$27.48 & 27.49±0.76 & \textbf{$^\dagger$54.41} & 54.43±0.60 \\
 & CTC+Reorder & $^\dagger$29.68 & 29.68±0.61 & $^\dagger$23.99 & 23.98±0.38 & $^\dagger$23.90 & 23.91±0.72 & $^\dagger$51.41 & 51.44±0.57 \\
 & CTC+ASN & \textbf{$^\dagger$40.34} & 40.33±0.50 & \textbf{$^\dagger$28.81} & 28.81±0.36 & $^\dagger$27.43 & 27.45±0.75 & $^\dagger$54.24 & 54.27±0.57 \\ \midrule
\multirow{7}{*}{$k=7$} & wait-$k$ & 40.78 & 40.76±0.67 & 29.50 & 29.50±0.48 & \textbf{30.28} & 30.32±0.80 & \textbf{56.44} & 56.47±0.62 \\
 & wait-$k$+Pseudo & \textbf{*42.34} & 42.34±0.62 & \textbf{*30.50} & 30.50±0.45 & \textbf{30.53} & 30.56±0.82 & \textbf{56.47} & 56.49±0.64 \\
 & wait-$k$+Reorder & *40.23 & 40.23±0.61 & *29.03 & 29.03±0.45 & *28.77 & 28.79±0.75 & *55.55 & 55.58±0.57 \\
 & CTC & 34.14 & 34.12±0.58 & 25.96 & 25.95±0.40 & 26.77 & 26.78±0.72 & 53.82 & 53.84±0.62 \\
 & CTC+Pseudo & $^\dagger$36.04 & 36.04±0.63 & $^\dagger$27.27 & 27.27±0.41 & $^\dagger$27.66 & 27.67±0.75 & $^\dagger$54.70 & 54.72±0.58 \\
 & CTC+Reorder & $^\dagger$29.45 & 29.44±0.64 & $^\dagger$23.86 & 23.85±0.40 & $^\dagger$24.21 & 24.23±0.70 & $^\dagger$51.50 & 51.53±0.57 \\
 & CTC+ASN & $^\dagger$40.81 & 40.80±0.49 & $^\dagger$29.22 & 29.21±0.35 & $^\dagger$27.30 & 27.32±0.74 & $^\dagger$54.18 & 54.21±0.57 \\ \midrule
\multirow{7}{*}{$k=9$} & wait-$k$ & 43.80 & 43.79±0.63 & 31.42 & 31.42±0.45 & 30.52 & 30.55±0.77 & 56.77 & 56.79±0.61 \\
 & wait-$k$+Pseudo & \textbf{*44.99} & 44.98±0.57 & \textbf{*32.23} & 32.23±0.45 & \textbf{*30.99} & 31.02±0.79 & \textbf{*57.14} & 57.16±0.62 \\
 & wait-$k$+Reorder & *43.27 & 43.27±0.62 & *30.92 & 30.92±0.44 & *29.37 & 29.39±0.80 & *56.25 & 56.27±0.58 \\
 & CTC & 34.20 & 34.18±0.60 & 26.03 & 26.02±0.41 & 27.37 & 27.38±0.74 & 54.37 & 54.39±0.59 \\
 & CTC+Pseudo & $^\dagger$36.83 & 36.83±0.64 & $^\dagger$27.67 & 27.66±0.41 & $^\dagger$27.72 & 27.74±0.75 & $^\dagger$54.75 & 54.77±0.58 \\
 & CTC+Reorder & $^\dagger$29.81 & 29.79±0.65 & $^\dagger$24.07 & 24.06±0.40 & $^\dagger$24.32 & 24.33±0.71 & $^\dagger$51.66 & 51.68±0.58 \\
 & CTC+ASN & $^\dagger$40.83 & 40.82±0.51 & $^\dagger$29.21 & 29.20±0.35 & $^\dagger$28.00 & 28.02±0.78 & $^\dagger$54.71 & 54.74±0.60 \\ \bottomrule
\end{tabular}}
\caption{Detailed quality metrics statistics on both datasets. Significance tests are conducted with paired bootstrap resampling. ``*'' suggests \textbf{significantly different (better or worst)} from the wait-$k$ baseline with $p$-value $<0.05$. ``$\dagger$'' suggests significantly different from the CTC baseline. \textbf{Bold} text suggests the best value in the same $k$. If multiple values are in bold, it means that these values are not significantly different according to paired bootstrap resampling.}
\label{tab:detailed_results}
\end{table*}

